%% file: main.tex
\pdfoutput=1

\documentclass[11pt]{article}

\usepackage[final]{acl}

\usepackage{times}
\usepackage{latexsym}

\usepackage[T1]{fontenc}

\usepackage[utf8]{inputenc}

\usepackage{microtype}

\usepackage{inconsolata}

\usepackage{graphicx}

\usepackage{multirow}
\usepackage{enumitem}

\usepackage[greek,english]{babel}

\usepackage{float}

\title{Can LLMs Solve and Generate Linguistic Olympiad Puzzles?}

\author{Neh Majmudar \\
  CUNY \\
  \texttt{nmajmudar@gradcenter.cuny.edu} \\\And
  Elena Filatova \\
  CUNY \\
  \texttt{efilatova@citytech.cuny.edu} \\}

\begin{document}
\maketitle
\begin{abstract}

In this paper, we introduce a combination of novel and exciting tasks: the solution and generation of linguistic puzzles. We focus on puzzles used in Linguistic Olympiads for high school students. We first extend the existing benchmark for the task of solving linguistic puzzles. We explore the use of Large Language Models (LLMs), including recent state-of-the-art models such as OpenAI's o1, for solving linguistic puzzles, analyzing their performance across various linguistic topics. 
We demonstrate that LLMs outperform humans on most puzzles types, except for those centered on writing systems, and for the understudied languages. We use the insights from puzzle-solving experiments to direct the novel task of puzzle generation. We believe that automating puzzle generation, even for relatively simple puzzles, holds promise for expanding interest in linguistics and introducing the field to a broader audience. This finding highlights the importance of linguistic puzzle generation as a research task: such puzzles can not only promote linguistics but also support the dissemination of knowledge about rare and understudied languages.
  




\end{abstract}

\section{Introduction}
\input{introduction}

\section{Related Work}
\input{related_work}

\section{Linguistic Puzzles Collection}
\input{dataSet}

\section{Using LLMs to Solve Linguistic Puzzles}
\input{people_vs_llms}

\section{Linguistic Puzzles Generation}
\input{problems_theory}

\section{Conclusion}
\input{conclusion}
\clearpage
\section{Limitations}
\input{limitations}

\section{Acknowledgments}
\input{acknowledgments}

\bibliography{custom}
\newpage
\appendix

\section{Checklist}
\input{checklistResNLP}

\section{Appendix A: Examples of the UKLO Linguistic Puzzles}
\input{appendix_a}
\clearpage
\section{Appendix B: Examples of the Generated Linguistic Puzzles}
\input{appendix_b}

\end{document}

%% file: introduction.tex
\label{sec:Intro}

Large Language Models (LLMs) are used for both technical and creative tasks. In this work, we investigate LLMs' ability to generate and solve linguistic puzzles designed for high school-level competitions, such as the International Linguistics Olympiad (IOL)\footnote{\url{https://ioling.org/}} and national contests. We argue that studying linguistic puzzles informs our understanding of both the technical capabilities and creative potential of LLMs.


Solving linguistic puzzles combines logical thinking as well as a creative approach to problem-solving. According to the IOL's site: `\textit{The competition challenges participants to analyze the grammar, structure, culture, and history of different languages and to demonstrate their linguistic abilities through puzzles and problem-solving challenges.}'' 


The IOL and several national Linguistic Olympiads make their puzzles publicly available for future participants to practice. Prior work has attempted to analyze the complexity of linguistic puzzle-solving task~\cite{10.5555/1627306.1627321,bozhanov-derzhanski-2013-rosetta,sahin-etal-2020-puzzling}. 

The puzzle generation process is creative and exciting but also tedious, often requiring the expertise of highly skilled linguists to ensure validity. This challenge is compounded by the lack of formal criteria for evaluating the quality of linguistic puzzles. In our project, we build on the work of~\cite{Gleason,Zaliznyak,WordLetterNumber} to develop formal criteria that can serve as a foundation for automatic linguistic puzzle generation. While linguistic puzzle generation is an exciting task in its own right, advancing generation methods offers practical benefits for educational outreach by enabling the rapid creation of puzzles of varying difficulty and thereby encouraging broader engagement with linguistic studies.



Before proceeding to the puzzle generation process, we describe existing the collections of linguistic puzzles. In Section~\ref{sec:dataSet}, we present the \textsc{LingOly} benchmark~\cite{LingOly}, which consists of puzzles created for the United Kingdom Linguistics Olympiad (UKLO).\footnote{\url{https://www.uklo.org/}} \textsc{LingOly} spans six linguistic topics: phonology, morphology, syntax, semantics, number systems, and compound problems. Additionally, we introduce a supplementary set of puzzles focusing on various writing systems.

To better understand the nature of linguistic puzzles, we examine the puzzle solving process. In Section~\ref{sec:peopleVsLLM}, we present results from applying LLMs (with and without explicit reasoning capabilities) to puzzles across a range of linguistic topics. Our evaluation shows that newer, reasoning-enabled LLMs frequently outperform general-purpose LLMs. Furthermore, both types of LLMs outperform human solvers in most linguistic topics, with the notable exception of puzzles focused on writing systems. This finding enables a deeper investigation into the reasoning capabilities and limitations of LLMs.

In Section~\ref{sec:theorySection}, we describe our attempt to incorporate the principles from the theory of linguistic puzzle design into LLM prompts for the purpose of generating new puzzles. We incorporate the insights from the puzzle solving experiment into the puzzle generation task. We conduct a series of experiments in which LLMs are tasked with the novel challenge of linguistic puzzle \textbf{generation}. Creating high-quality puzzles requires a blend of expertise, scientific insight, and creativity. Evaluating the quality of generated puzzles is a non-trivial task, as only a small number of linguists have experience in puzzle design. Since the generated puzzles are intended for use in linguistic Olympiads, we rely on input from linguistics Olympiad participants to help develop the evaluation procedure.

%% file: related_work.tex
\label{relatedWork}

LLMs have demonstrated efficiency across a variety of tasks~\cite{minaee2024large}. For text-related tasks, such as understanding and analysis, generation and transformation, and conversational tasks, LLMs often outperform traditional pre-trained language models~\cite{zhou2024comprehensive}. Pre-trained on diverse text data, LLMs have proven successful in solving problems such as SQL query generation~\cite{10542806}, software testing~\cite{10391027}, and mathematical problem-solving~\cite{MathEDU}. Additionally, LLMs are effectively used for creative tasks, including short story writing~\cite{10.1145/3490099.3511105} and text adjustment based on user preferences~\cite{Ouyang2022TrainingLM}.

OpenAI claims that their o1 model that includes reasoning capabilities ``\textit{ranks in the 89-th percentile on competitive programming questions (Codeforces), places among the top 500 students in the US in a qualifier for the USA Math Olympiad (AIME), and exceeds human PhD-level accuracy on a benchmark of physics, biology, and chemistry problems (GPQA).}''\footnote{\url{https://openai.com/index/learning-to-reason-with-llms/}} However, when using a different benchmark for Math Olympiad problems, namely 2025 USAMO\footnote{\url{https://artofproblemsolving.com/wiki/index.php/United_States_of_America_Mathematical_Olympiad}} problems, Petrov at el.~\shortcite{petrov2025proofbluffevaluatingllms} claim that ``\textit{current LLMs are inadequate for rigorous mathematical reasoning tasks, highlighting the need for substantial improvements in reasoning and proof generation capabilities.}''

Giadikiaroglou et al.~\shortcite{giadikiaroglou2024puzzle} provide a survey for puzzle solving approaches that use LLMs' reasoning. According to this survey, while LLMs excel at generating human-like text, they often struggle with complex logical puzzles requiring advanced inference and multi-step reasoning. Linguistics puzzles are \textbf{not} analyzed within this survey. 

LLMs are successfully used for question generation given a short story~\cite{yao-etal-2022-ais} or given a query path in the knowledge graph constructed from the input text~\cite{wang-etal-2020-pathqg}. Both methodologies are evaluated using a gold standard human-generated set of questions against which the generated questions are compared.

In our work, we focus on linguistic puzzles designed for Linguistic Olympiads~\cite{10.5555/1627306.1627321}. Most of these puzzles fall into two types: Rosetta Stone and Match-up. Rosetta Stone puzzles are typically bilingual and consist of sets of corresponding words or phrases from different languages or writing systems, with most correspondences explicitly provided. The Xhosa puzzle (App.~\ref{sec:AppA}, Fig.~\ref{fig:Xhosa}) is an example of a Rosetta Stone puzzle. \citet{sahin-etal-2020-puzzling} apply various methods to automatically solve Rosetta Stone-type linguistic puzzles. Match-up puzzles feature sets of words or phrases in multiple languages or writing systems without given correspondences; participants must infer the mappings themselves. The Waama puzzle (App.~\ref{sec:AppA}, Fig.~\ref{fig:Waama}) illustrates this type. 

%% file: dataset.tex
\label{sec:dataSet}

\subsection{UKLO Puzzles in \textsc{LingOly} Dataset}

For our initial experiments, we use a subset of the UKLO linguistic puzzles\footnote{\url{https://www.uklo.org/past-exam-papers/}} assembled into the \mbox{\textsc{LingOly}} benchmark~\cite{LingOly}. While there are other linguistics puzzles datasets~\cite{sahin-etal-2020-puzzling,chi-etal-2024-modeling}, and many national linguistic competition post their puzzles and solutions online, the UKLO organizers, in addition to the puzzles and their solutions, list several attributes describing their puzzles. These attributes include: puzzle difficulty, linguistic topic (writing system, morphology, etc.), question format (Rosetta Stone, Match-up, etc.), language family, and other attributes. \citet{LingOly} describe the application of LLMs to solving the puzzles from the \mbox{\textsc{LingOly}} benchmark and show that LLMs outperform humans on several types of linguistic puzzles, however they also notice: ``\textit{in absence of memorisation, true multi-step out-of-domain reasoning remains a challenge for current language models}.''


Currently, UKLO lists 220 puzzles for the competitions held between 2010 and 2024. \textsc{LingOly} contains 90 out of these 220 puzzles. Each puzzle contains ``\textit{a preamble, which gives general background on the language in question; a context, which provides required background to solve the puzzle, such as example translations; and questions, which are sometimes further divided into subquestions.}'' Most UKLO puzzles contain several questions. App.~\ref{sec:AppA}, Fig.~\ref{fig:Warlpiri} contains the problem 2024 UKLO puzzle regarding the Warlpiri language. This puzzle contains two questions, each of which has subquestions (problems). \textsc{LingOly} contains 1,133 problems for 90 UKLO puzzles. 

\textsc{LingOly} contains UKLO puzzles of \textbf{five difficulty levels} (from easiest to most difficult): Breakthrough (Br), Foundation (Fn), Intermediate (Int), Advanced (Adv), and Round\_2 (R2). The \textbf{six linguistic topics} covered in \mbox{\textsc{LingOly}} are: Phonology (Ph), Semantics (Se), Morphology (Mo), Numbers (Nu), Compounding (Co), and Syntax (Sy).\footnote{In the charts and tables presented in this paper, we use the listed abbreviations when referring to difficulty and topic.} Also, each UKLO puzzle has information about the corresponding score (percent) that indicates the average participants' scores on the problem. ``\textit{A high score of 90\% indicates that, on average, students scored 90\% on that particular question}''.\footnote{\url{https://www.uklo.org/technical-information}} If a puzzle is cross-listed for different difficulty levels, a separate score is provided for each of the difficulty levels. The percentage scores are normalized as different puzzles have different maximum scores. Puzzle questions can consist of several parts. For example, the 2024 Warlpiri puzzle (App.~\ref{sec:AppA}, Fig.~\ref{fig:Warlpiri}) consists of two questions with a combined possible score of 5 points. The 2021 Waama puzzle (App.~\ref{sec:AppA}, Fig.~\ref{fig:Waama}) contains one question with a maximum possible score of 10 points. The answers provided by UKLO contain the point distributions for the solutions. We use these point distributions to evaluate the ability of \mbox{OpenAI's o1} to solve puzzles.  

Table~\ref{tab:LingOlyCorpusStatistics} contains the distribution of the \mbox{\textsc{LingOly}} puzzles across two dimensions: linguistic topic and difficulty. Table~\ref{tab:LingOlyCorpusStatistics} contains the number of puzzles, rather than the combined number of questions for all the puzzles. According to this table, the dataset contains no Compounding puzzles at the Breakthrough or Foundation levels. Several puzzles are used for two groups of participants, and thus, have two levels of difficulty, each of which has a separate average score assigned to them. Also, several puzzles cover more than one linguistic topic. For example, the Warlpiri puzzle (App.~\ref{sec:AppA}, Fig.~\ref{fig:Warlpiri}) has two difficulty scores (its Breakthrough score is 41\% and its Foundation score is 45\%); and it covers two linguistic topics: morphology and phonology. Such puzzles are counted several times in Table~\ref{tab:LingOlyCorpusStatistics}: once for each difficulty level/linguistic topic.

\begin{table}
\centering
\begin{tabular}{|c|c|c|c|c|c|c|}
\hline
     & {\textbf{Ph}} & {\textbf{Se}} & {\textbf{Mo}} &{\textbf{Nu}} & {\textbf{Co}} & {\textbf{Sy}} \\
\hline
{\textbf{Br}} & 7 & 1 & 7 & 1 & 0 & 3 \\
\hline
{\textbf{Fn}} & 10 & 4 & 16 & 1 & 0 & 11 \\
\hline
{\textbf{Int}} & 6 & 4 & 15 & 1 & 1 & 8 \\
\hline
{\textbf{Adv}} & 9 & 4 & 18 & 4 & 2 & 7 \\
\hline
{\textbf{R2}} & 8 & 6 & 13 & 2 & 2 & 13 \\
\hline
\end{tabular}
\caption{\textbf{Distribution of the \textsc{LingOly} puzzles} across six linguistic topics and five difficulty dimensions. The linguistic topics are: Phonology (Ph), Semantics (Se), Morphology (Mo), Numbers (Nu), Compounding (Co), and Syntax (Sy). The difficulty dimensions are: Breakthrough (Br), Foundation (Fn), Intermediate (Int), Advanced (Adv), and Round\_2 (R2).}
\label{tab:LingOlyCorpusStatistics}
\end{table}

\subsection{UKLO Writing Systems Puzzles}
\label{sec:DataWritingSystems}

In this work, in addition to the \textsc{LingOly} puzzles, we use the UKLO puzzles that focus on deciphering writing systems. The UKLO website lists 41 such puzzles, five of which combine writing systems with another linguistic topic. Among the 36 puzzles that focus solely on writing systems, five lack participants' performance data. Therefore, in this project, we use the remaining 31 puzzles, which exclusively focus on writing systems and include participant performance scores for evaluation.

The UKLO puzzles that deal with writing systems contain a variety of inscriptions, symbols, or images as questions (App.~\ref{sec:AppA}, Figs.~\ref{fig:Ditema},~\ref{fig:Georgian}). These puzzles cannot be parsed into a text format that is used in \textsc{LingOly}. Thus, we split these puzzles into 2 PDF files: one~--~for the puzzle preamble, context, and the questions associated with this puzzle, and the other one~--~with the answer key, solution, grading instructions, and the answers explanation. Each page of the first PDF file (puzzle preamble, context, and questions) is converted into image files. These image files are submitted to LLMs. 


%% file: people_vs_llms.tex
\label{sec:peopleVsLLM}

\begin{table*}[t]
\centering
\begin{tabular}{|c|ccc|ccc|ccc|ccc|ccc|ccc|}
\hline
\multirow{2}{*}{} & \multicolumn{3}{c|}{\textbf{Ph}} & \multicolumn{3}{c|}{\textbf{Se}} & \multicolumn{3}{c|}{\textbf{Mo}} & \multicolumn{3}{c|}{\textbf{Nu}} & \multicolumn{3}{c|}{\textbf{Co}} & \multicolumn{3}{c|}{\textbf{Sy}} \\
\cline{2-19}
 & \textbf{H} &      \textbf{C} & \textbf{O} & \textbf{H} &      \textbf{C} & \textbf{O} & \textbf{H} &      \textbf{C} & \textbf{O} & \textbf{H} &      \textbf{C} & \textbf{O} & \textbf{H} &      \textbf{C} & \textbf{O} & \textbf{H} &      \textbf{C} & \textbf{O} \\
\hline
\textbf{Br} & 50 &  74 & \textbf{88} & 69  & - & \textbf{91} & 44 & \textbf{92} & 89 & 78  & 92 & \textbf{100} &  * & * & * & 46  & - & \textbf{98} \\
\hline
\textbf{Fn} & 54 &  80 & \textbf{82} & 46 & 77 & \textbf{81} & 47 &  46 & \textbf{71} & 41  & - & \textbf{100} & *  & * & * & 53 & 81 & 81 \\
\hline
\textbf{Int} & 57  & 45 & \textbf{69} & 37  & 44 & \textbf{57} & 54  & 45 & \textbf{67} & \textbf{22}  & - & 0 & 47 & - & \textbf{100} & 61 & 55 & \textbf{76} \\
\hline
\textbf{Adv} & 45 &  58 & \textbf{68} & 31  & 26 & \textbf{53} & 48  & 50 & \textbf{67} & 18 & 8 & \textbf{26} & 32  & 42 & \textbf{65} & 42  & 59 & \textbf{66} \\
\hline
\textbf{R2} & \textbf{37} & 25 & 31 & 33 & 42 & \textbf{58} & 44 & 25 & \textbf{49} & 16 & 16 & \textbf{50} & 16 & \textbf{24} & 2 & 47 & 30 & \textbf{51} \\
\hline
\end{tabular}
\caption{\textbf{Average Scores by Linguistic Topic and Difficulty Level on the \mbox{\textsc{LingOly}} Benchmark}.
\\
\textbf{H} - The average human performance reported on the UKLO website; 
\textbf{C} - The best exact match scores of the \textit{Claude Opus} model reported by Bean et al.~\shortcite{LingOly}; \textbf{O} - The exact match score for the OpenAI o1.
\\
`*` corresponds to $0$ in Table~\ref{tab:LingOlyCorpusStatistics}: there are no \mbox{\textsc{LingOly}} puzzles of this type. `-` corresponds to the cases where LLM does not produce a result for the linguistic puzzle of the corresponding linguistic topic/difficulty level. 
\\
The linguistics topic and difficulty abbreviations are the same as in Table~\ref{tab:LingOlyCorpusStatistics}.}
\label{tab:peopleVsLLMsv2}
\end{table*}

\subsection{Experiments on the \mbox{\textsc{LingOly}} dataset}

\citet{LingOly} use $11$ state-of-the-art general-purpose LLMs to solve \mbox{\textsc{LingOly}} puzzles. These LLMs are: Llama 3 8B and 70B~\cite{dubey2024llama}, Mixtral 8x7B~\cite{jiang2024mixtral}, Aya 23 35B~\cite{aryabumi2024aya}, Gemma 7B~\cite{team2024gemma}, Llama 2 70B~\cite{touvron2023llama}, GPT-4o~\cite{hurst2024gpt}, GPT-4~\cite{achiam2023gpt}, GPT-3.5~\cite{brown2020language}, Claude Opus~\cite{TheC3}, Gemini 1.5 Pro~\cite{team2024gemini}, and Command R+~\cite{CohereR}. 

For our experiments, we use OpenAI's o1.\footnote{\url{https://cdn.openai.com/o1-system-card-20241205.pdf}} We aim to investigate if the reasoning capabilities of OpenAI's o1 enhance the puzzle solving performance. We evaluate the performance of OpenAI's o1 ability to solve linguistic puzzles by using the actual scoring instructions listed on the UKLO puzzle sheets. We use the \mbox{\textsc{LingOly}} benchmark to compare the ability of OpenAI's o1 (LLM \textit{with reasoning}) to solve linguistic puzzles and compare our results with the results for other LLMs. 





The UKLO website reports one performance score per puzzle, without splitting this score per question. \citet{LingOly} report one average score across all the questions for all the puzzles of a particular topic/difficulty level pair. When running OpenAI's o1 we use the \textbf{exact match} evaluation metric and average OpenAI's o1 scores computed for a particular topic/difficulty level pair. The \textbf{exact match} metric counts only the exact answers corresponding to the exhaustive UKLO answer. Based on the results reported by \citet{LingOly}, the model that produces the best exact match results is Claude Opus.

As per Table~\ref{tab:LingOlyCorpusStatistics}, \mbox{\textsc{LingOly}} does not contain \textit{Beginner} and \textit{Foundation} puzzles for the \textit{Compounding} topic. In several cases, LLMs do not produce any results. Often, these are the cases when there is only one puzzle of a particular linguistic topic/difficulty level pair (see the \textit{Numbers} topic for \textit{Beginner}, \textit{Foundation}, and \textit{Intermediate} difficulty).  

Table~\ref{tab:peopleVsLLMsv2} contains the results for human participants based on the scores provided by the UKLO website (H), the best exact match results by Claude Opus (C); and the exact match results that we get by running \mbox{OpenAI's o1} LLM with the reasoning capability (O). Like in Table~\ref{tab:LingOlyCorpusStatistics}, we analyze the distribution of the \textsc{LingOly} questions across six linguistic topics and five difficulty dimensions. The linguistic topics are: Phonology (Ph), Semantics (Se), Morphology (Mo), Numbers (Nu), Compounding (Co), and Syntax (Sy). The difficulty dimensions are: Breakthrough (Br), Foundation (Fn), Intermediate (Int), Advanced (Adv), and Round\_2 (R2). All the presented scores are average scores computed for topic/difficulty level pairs across the puzzles used in \mbox{\textsc{LingOly}}. Following the \mbox{\textsc{LingOly}} notation, the average numbers are integers. We round all the numbers (average human performance and average OpenAI's o1 performance) down to integers using the floor function. Table~\ref{tab:peopleVsLLMsv2} compares the performance of \mbox{OpenAI's o1} with the previously reported results for Claude Opus. We observe improvements in several categories, though performance remains mixed across different topics and difficulty levels.


\subsection{Performance Analysis for OpenAI's o1 \mbox{\textsc{LingOly}} Puzzles}

Out of the 19 puzzles for which OpenAI's o1 provides 100\% correct solution, only 3 puzzles are of \textit{Advanced} difficulty level and 1 puzzle is from \textit{Round 2}, which is the most difficult level. The rest of the correctly solved puzzles are from lower difficulty levels. The languages on which the reasoning model does well are primarily those that are well-known and have vast resources, e.g. \textit{Italian}, 
\textit{Japanese}, 
\textit{Turkish}, 
\textit{Finnish}, etc. 
We believe that perfect scores are achieved based on the LLMs' access to vast corpora for these languages. Thus, the question arises if LLMs (both with and without reasoning) \textit{solve} linguistic puzzles, or merely provide translations based on their \textit{knowledge} of the language used in the puzzle without even attempting to solve the puzzles based on the context provided on the puzzle sheet. 

According to our observation, LLMs (including OpenAI's o1) do not perform well on the puzzles that require deep puzzle context understanding. For example, for the Maonan puzzle (App.~\ref{sec:AppA}, Fig.~\ref{fig:Maonan})  OpenAI's o1 gets 0\%. The puzzle's context contains clues about the use of different words for male/female. Using this information is necessary for solving the puzzle. Thus, we conclude: OpenAI’s o1 cannot fully use its reasoning capabilities within unfamiliar settings. Also, LLMs perform badly on the puzzles based on the poor-resourced languages: Wik-Mungkan (App.~\ref{sec:AppA}, Fig.~\ref{fig:WikMungkan}) is spoken by 1,650 speakers; Ngkolmpu (App.~\ref{sec:AppA}, Fig.~\ref{fig:Ngkolmpu}) is spoken by about a hundred people.

Four UKLO puzzles are generated for Constructed Language: Afrihili, Blazon, Esperanto, Centauri and Arcturan. Centuri and Arcutan are generated specifically for a UKLO puzzle; Esperanto and Afrihili are well-documented attempts to create Pan-European and Pan-African languages with regular grammar. Out of these four puzzles, only the Afrihili puzzle is used in the \mbox{\textsc{LingOly}} corpus. This puzzle is a Rosetta Stone puzzle dealing with Morphology and Semantics used for Round 2 in 2019; human performance is 89\%, Claude's and OpenAI's o1 performances are 31\% and 48\% respectively. Afrihili does not have a lot of texts written in it and is not well-studied. Thus, it can be treated as a poor-resourced language.

For the Match-Up puzzles, where OpenAI's o1 fails to come up with an answer, the output is often organized in perfect alphabetical (or numeric) order. During the evaluation, we assign 0 to such ordered answers produced by OpenAI's o1, even if some answers are accidentally matched correctly. This situation occurs in five puzzles. The difficulty levels for these puzzles are: two puzzles of Round~2~(App.~\ref{sec:AppA}, Figs.~\ref{fig:WikMungkan},~\ref{fig:Maonan}); two puzzles of the Advanced (App.~\ref{sec:AppA}, Figs.~\ref{fig:Ngkolmpu},~\ref{fig:Mazateco}); and one puzzles of Foundation/Intermediate level (App.~\ref{sec:AppA}, Figs.~\ref{fig:Maltese}).

\subsection{Experiments on the Linguistic Puzzles Dealing with Writing Systems}

As stated in Section~\ref{sec:DataWritingSystems}, in our work, we use an additional linguistic topic that is not covered in the \mbox{\textsc{LingOly}} benchmark: Writing Systems. 
Puzzles on Writing Systems explore language representation through written symbols or scripts and examine how languages are visually encoded and how writing conventions function. 

To solve 31 UKLO puzzles that are centered solely around writing systems we use OpenAI's o1 and one of the models without reasoning, GPT-4o. GPT-4o is among the 11 LLMs used by \citet{LingOly} and is the second-best performing model losing only to Claude Opus. We do not use the best-performing Claude Opus due to its output token length limit, which occasionally results in the LLM not solving all the questions in the puzzle.

Table~\ref{tab:peopleVsLlmsWritingSys} contains information about the number of UKLO Writing System puzzles split by the difficulty score; the average percentage scores by participants, GPT-4o, and OpenAI's o1. On average, OpenAI's o1 outperforms GPT-4o. Out of 31 writing systems puzzles, OpenAI's o1 outperforms GPT-4o in 9 cases, while GPT-4o outperforms OpenAI's o1 in 4 cases. Moreover, humans outperform both LLMs on difficult puzzles.

\begin{table}
    \centering
    \begin{tabular}{|c||c||c|c|c|}
        \hline
         & \textbf{\# of Puzzles} & \textbf{H} & \textbf{4o} & \textbf{o1}  \\
        \hline
       \textbf{Br} & 8 & 47.5 & 48.5 & \textbf{55.9}  \\
        \textbf{Fn} & 12 & 51.3 & 49.4 & \textbf{55.4}  \\
        \textbf{Int} & 13 & \textbf{45.8} & 40.7 & 42.3  \\
        \textbf{Adv} & 12 & \textbf{27.6}  & 21.6 & 22.9 \\
        \textbf{R2} & 5 & \textbf{45.2} & 15.6 & 24.5  \\
        \hline
    \end{tabular}
    \caption{\textbf{Comparison of Scores for the Writing System Puzzles by Difficulty Level}. \underline{H} - The average human performance reported on the UKLO website; \underline{4o} - The exact match score for the GPT-4o on the Writing System puzzles;
    \underline{o1} - The exact match score for the \mbox{OpenAI's o1} on the Writing System puzzles.
    \\
    The difficulty abbreviations are the same as in Table~\ref{tab:LingOlyCorpusStatistics}.}
    \label{tab:peopleVsLlmsWritingSys}
\end{table}

\subsection{Performance Analysis for GPT-4o and OpenAI's o1 on the UKLO Writing System Puzzles}

For the hardest problems (three highest difficulty levels) people \textbf{do} outperform LLMs.

When analyzing the solutions provided by both GPT-4o and OpenAI's o1, we confirm our hypothesis from the previous section: whenever possible, LLMs rely on their knowledge of the language rather than make inferences based on the puzzle context. For example, one of the 2015 puzzles involves the Georgian alphabet (App.~\ref{sec:AppA}, Fig.~\ref{fig:Georgian}). In this puzzle, participants must match location names written in Georgian with their English equivalents. To do it participants should match Georgian letters with their Latin (English) counterparts.
GPT-4o correctly performs this matching and, for the Georgian word \includegraphics[scale=0.175]{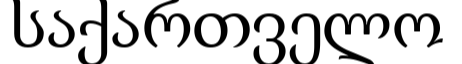}, produces the expected answer: \textit{Sakartvelo}. In contrast, OpenAI's o1 outputs \textit{Georgia}. While \textit{Georgia} is technically correct—since \textit{Sakartvelo} is the Georgian name for the country of \textit{Georgia}\footnote{\url{https://en.wikipedia.org/wiki/Georgia_(country)}}—it is not the answer that can be deduced from the puzzle context, nor the one intended by the puzzle's authors. Given that GPT-4o produced the expected answer, we hypothesize that OpenAI's o1 initially arrived at \textit{Sakartvelo} but then leveraged its knowledge of Georgian and converted it to \textit{Georgia}. Notably, both models answered the remaining questions in this puzzle correctly. Thus, when solving linguistic puzzles, OpenAI's o1 does not rely solely on the puzzle context but rather incorporates its broader knowledge of the language.

To test the hypothesis that whenever possible LLMs rely on their knowledge of the language run an additional experiment: we create a new puzzle for the Greek alphabet following the 2015 Georgian alphabet puzzle structure.
This Greek puzzle (App~\ref{sec:AppA}, Tbl.~\ref{tab:GreekPuzzle})
has a Rosetta Stone-style context where Greek locations, written in all capital letters, are listed with their translations. The task is to translate the Greek word \foreignlanguage{greek}{ΕΛΛΑΔΑ}. We use capital letters for Greek words in this puzzle to avoid using the notation for stress that is mandatory for the Greek words written in small letters. The answer provided by OpenAI's o1 is the following: `` Ell{\'a}da (the modern Greek word for Greece).'' While in contrast to the Georgian example, the LLM produces the correct answer, the presence of the explanation that \textit{Ell{\'a}da} can be used for the name of the country instead of \textit{Greece} clearly demonstrates that answer is obtained given the knowledge of the Greek language rather than purely deduced from the puzzle context. Moreover, the provided answer contains the information about the stressed syllable, however, the puzzle context does not contain any examples of stress for either of the languages. 

%% file: problems_theory.tex
\label{sec:theorySection}

In this section, we discuss the task of linguistic puzzle generation using LLMs. To the best of our knowledge, this is the first attempt to automatically generate Olympiad-level linguistic puzzles. 

Generating interesting puzzles for linguistic competitions is a challenging task. 
Linguistic puzzles used in linguistic competitions typically require multi-step reasoning over the limited data presented in the puzzle. Moreover, 
the puzzle statement should contain all the information necessary for puzzle solving. This requirement for linguistic puzzles goes beyond deep understanding of a human language as the puzzle generation task implies that reasoning is needed to solve the output puzzle.

In this work, we demonstrate that current state-of-the-art LLMs can generate puzzles that are not necessarily on the Olympiad-level, but can be used for smaller, preliminary competitions, or for providing an easy starting point for those who see such linguistic puzzles for the first time.

The generation puzzles generation procedure described in this section draws insights from the puzzle-solving experiment described in Section~\ref{sec:peopleVsLLM}. Specifically, the generated puzzles are designed to challenge students' genuine reasoning and pattern detection, minimizing reliance on external language knowledge.

Before proceeding to the experiment where we apply LLMs to linguistic puzzle generation, we first describe the theory behind what constitutes a good linguistic puzzle. While puzzle generation is undoubtedly a creative task, formal rules should be applied to assess the generated puzzle. In this work, we focus solely on evaluating whether the generated linguistic puzzles are valid or not. We do not assess their creativity.

\subsection{Theory of Linguistic Puzzles}
\label{sec:theory}

Since 1965, annual competitions for high school students focused on solving linguistic puzzles have been held in Moscow. The first collections of self-contained linguistic puzzles are described in~\cite{Gleason, Zaliznyak}. One key feature of these puzzles is that no external knowledge is required to solve them.

Alfred Zhurinsky is one of the founders of linguistic competitions. According to \citet{WordLetterNumber}, when considering what makes a good linguistic puzzle, linguists should refer to research on Gestalt Psychology. Based on this research, the important characteristics of linguistics puzzles are:

--~accessible solution;           

 --~self-contained nature of the puzzle statement;       

 --~the puzzle should be meaningful according to the solver's life experience;    

 --~there should be multiple ways to approach the puzzle solution where only one of those approaches leads to the correct solution.   
%
%
%
%

Zhurinsky was among the first to not only define the characteristics of a linguistic puzzle suitable for competition but also to describe three criteria for eliminating linguistic puzzles that are \textbf{not} valid: 

\noindent (1) the puzzle is formulated in a way that it contains parasitic solutions: logically plausible solutions that are incorrect given the language for which the puzzle is created; 

\noindent (2) the description of the linguistic phenomenon discovered as part of the puzzle solution contains inconsistencies or lacks clarity; 

\noindent (3) the puzzle solution cannot be described by the material available in the puzzle context. 

The linguistic puzzles that can be invalidated based on the three criteria above should be avoided by the authors who create linguistic puzzles. Those puzzles that are used in the International and National Linguistics competitions are valid puzzles. 

\subsection{Linguistic Puzzles Generation}
\label{sec:puzzleGeneration}

Puzzle generation is a creative task. However, we focus on testing whether LLMs can generate \textit{valid} puzzles. Evaluating the creativity of the generated puzzles is beyond the scope of this work.

For puzzle generation, we use puzzles from \textsc{LingOly}, the Gestalt Psychology puzzle principles, and Zhurinsky's criteria for invalid puzzles. According to Table~\ref{tab:LingOlyCorpusStatistics}, \textsc{LingOly} contains the most questions for the morphology topic. Therefore, we focus on generating morphology puzzles. As training examples, we use four UKLO morphology puzzles from Rosetta Stone and Breakthrough-level categories that are part of \textsc{LingOly}. The generated puzzles should include not only questions but also their corresponding answers and explanations. To achieve this, we extend the \textsc{LingOly} puzzle sheets, which contain a preamble, context, and questions, by adding solutions and solution explanations.

We use GPT-4o and OpenAI's o1 LLMs to generate new morphology puzzles along with their solutions. The input generation process mirrors the one we used to evaluate the Writing System puzzles: we convert the UKLO puzzle files into images. In this experiment, in addition to the puzzle preamble, context, and questions, we also use the puzzle solutions and their corresponding explanations.

LLMs are tasked with generating the complete linguistic puzzle: preamble, context, questions, solutions, and explanations. We use two LLMs: GPT-4o and OpenAI's o1; and three settings:

\noindent\textbf{Zero Shot:} the prompt consists of Gestalt psychology principles and Zhurinsky's criteria, and tasks the LLM with creating similar puzzles;

\noindent\textbf{One Shot:} the prompt consists of Gestalt psychology principles, Zhurinsky's criteria, and one \textsc{LingOly} morphology puzzle to demonstrate the puzzle structure the LLM should generate. LLM's task is to generate similar puzzles;

\noindent\textbf{Few Shot:}  the prompt consists of Gestalt psychology principles, Zhurinsky's criteria, and four \textsc{LingOly} morphology puzzles as examples. LLM's task is to generate similar puzzles.

For all settings, the puzzles are written in English. Three languages that are the focus of the generated puzzles are Greek, Gujarati, and Spanish. The choice of languages is driven by the goal of testing the generation procedures across a diverse set of languages.
Two LLMs, GPT-4o and OpenAI's o1 are used for the puzzle generation. 

In total, we generate 18 puzzles (see Appendix~\ref{sec:AppB}). All these 18 puzzles follow the standard format: preamble (a short fact sheet about the language), context (Rosetta Stone examples used to deduce answers to the questions), questions, answers, and explanations. However, the puzzles generated using the \textbf{Zero Shot} setting, without an example puzzle, do not include the preamble and therefore lack a brief description of the puzzle language. 
 
For the \textbf{One Shot} setting, the example puzzle is the Lithuanian puzzle from UKLO 2018 (App.~\ref{sec:AppA}, Figs.~\ref{fig:LithuanianContext} and~\ref{fig:LithuanianSolution}). The structure of this puzzle's context is a conversation among friends. Thus, all puzzles generated for the \textbf{One Shot} setting are conversation among several friends. One generated puzzle (\mbox{OpenAI's o1} \textbf{Few Shot} Gujarati) contains a mistake: incorrect handling of Gujarati negation, and thus, is not a valid puzzle.
 
\subsection{Analysis of the Generated Puzzles}
\label{sec:PuzzlesError}


The task of linguistic puzzle generation is novel, and no standard evaluation procedure currently exists to assess the validity and quality of the generated puzzles. To design our evaluation framework, we relied on the expertise of three accomplished Linguistic Olympiad participants. Each expert was given five puzzles: two truncated UKLO puzzles (Q1.1, Swedish; App.~\ref{sec:AppA}, Fig.~\ref{fig:Swedish}, and Q2.1, Kabyle; App.~\ref{sec:AppA}, Fig.~\ref{fig:Kabyle}) and three automatically generated puzzles (GPT-4o~/~One-shot~/~Gujarati; OpenAI’s o1~/~One-shot~/~Greek; GPT-4o~/~Few-shot~/~Spanish). Of these five puzzles, only the Gujarati puzzle was written in non-Latin characters. The generated Greek puzzle submitted for evaluation was transliterated into Latin characters.


We asked our evaluators to attempt solving these five Rosetta Stone–type puzzles using a fill-in-the-blank (FITB) format. In addition, we asked our evaluators to indicate their confidence in the correctness of their solutions, estimate the difficulty level of each puzzle, and describe the features that made a puzzle easier or harder to solve. Evaluators were also asked to report their level of familiarity with the puzzle language. To ensure consistency, we requested that they spend no more than 15 minutes on each puzzle.

All evaluators solved the Swedish and Kabyle puzzles correctly. All evaluators have only cursory knowledge about the Swedish language structure and no knowledge of Kabyle. The Kabyle problem is labeled as \textit{beginner} level by all evaluators, while Swedish is labeled as \textit{beginner}-level by two, and \textit{intermediate}-level by one evaluator.

All evaluators solved the GPT-4o~/~Few-shot~/~Spanish puzzle correctly. All evaluators specified that they had a working knowledge of Spanish, and marked the puzzle as \textit{beginner} level.

All evaluators attempted to solve the OpenAI’s o1~/~One-shot~/~Greek puzzle. None of the evaluators had a prior knowledge of Greek, and thus, were not confident in the correctness of the solution. The evaluators labeled the puzzle as \textit{intermediate} or \textit{advanced}. 

Two evaluators who attempted the GPT-4o~/~One-shot~/~Gujarati puzzle, one of these evaluators provided correct solutions, while the other one provided incorrect solutions. Both evaluators expressed uncertainty and labeled the puzzle as \textit{advanced} due to their lack of Gujarati knowledge. The third evaluator did not attempt it, citing confusion over the inconsistent number of dialogue participants and whether this was significant.

According to our evaluators, the puzzle written in non-Latin scripts was perceived as more difficult (Gujarati) than the one written in Latin characters, even when the language itself normally uses a non-Latin script (Greek). This finding aligns with our observation that writing systems are the only linguistic topic in which LLMs perform worse than humans for the task of puzzle solving (Section~\ref{sec:peopleVsLLM}).

Table~\ref{tab:iaa_results} summarizes the inter-agreement among the answers submitted by our three evaluators. For the puzzle-specific fill-in-the-blanks questions, we calculate the performance for each of the evaluators and average the scores. For the Kabyle, Spanish and Swedish puzzles all the answers for all the evaluators were correct. The low score for the Greek puzzle is due to the fact that the puzzle could not be solved without external knowledge, and only one question by one evaluator was answered correctly. In the Gujarati case, 
one of the evaluators pointed out the confusion regarding if the number of the dialog participants (speaker(s) and interlocutor(s)) was significant. Thus, no answers were submitted by this evaluator. Out of the remaining two evaluators, one evaluator answered all the questions correctly, while the other one answered all the questions incorrectly. 

For the five feedback questions, we measure the agreement using Fleiss's Kappa~\cite{fleiss1971measuring} coefficient ($\kappa$).\footnote{Cohen's Kappa is used for the Gujarati case as only two evaluators submitted the answers.} This metric is appropriate for categorical data as it accounts for the probability of agreement occurring purely by chance, providing a more robust measure of reliability than simple percentage agreement.

\begin{table}[t]
\centering
\begin{tabular}{|c|c|c|c|}
\hline
\textbf{Language} & \textbf{Avg. FITB (\%)} & \textbf{Feedback ($\kappa$)} \\ \hline
\textbf{Greek} & 6.6 & 0.500 \\ \hline
\textbf{Gujarati} & 50 & 0.736 \\ \hline
\textbf{Kabyle} & 100 & 0.505 \\ \hline
\textbf{Spanish} & 100 & 0.638 \\ \hline
\textbf{Swedish} & 100 & 0.149 \\ \hline
\end{tabular}
\caption{\textbf{Inter-Annotator Agreement statistics} Average accuracy of puzzle-specific FITB questions, and Agreement of feedback questions across the five puzzles.}
\label{tab:iaa_results}
\end{table}

Following the evaluators' comments on the puzzle solving experience, we categorize the generated puzzles into four groups: puzzles that ask for the repetition of  context examples; puzzles that are invalid as they cannot be solved using only the information from the preamble and context; valid puzzles. Table~\ref{tab:language_cases} summarizes the distribution of the 18 generated puzzles across these four groups. 

 \begin{table}[t]
\centering
\begin{tabular}{|l|l|c|c|c|}
\hline
\textbf{Issue} & \textbf{Model} & \textbf{Greek} & \textbf{Gujarati} & \textbf{Spanish}  \\
\hline
\multirow{2}{*}{CR} & 4o & 1 & 1 & 1 \\
                    & o1 & f & - &   \\
\hline
\multirow{2}{*}{EK} & 4o & 0 & 0,f & 0 \\
                    & o1 & 1 & 1 & 1 \\
\hline
\multirow{2}{*}{VP} & 4o & f & - & f \\
                    & o1 & 0 & 0 & 0,f  \\
\hline
\multirow{2}{*}{IC} & 4o & - & - & - \\
                    & o1 & - & f & -  \\
\hline
\end{tabular}
\caption{\textbf{Categorization of issues in various settings for GPT-4o and OpenAI o1 in Gujarati, Spanish, Greek.} \underline{CR} - Context Repetition, \underline{EK} - External Knowledge is Required, \underline{VP} - Valid puzzle, \underline{IC} - Incorrect Context; \underline{0} - Zero-shot, \underline{1} - One-shot, \underline{f} - Few-shot}
\label{tab:language_cases}
\end{table}

\subsection{Context Repetition Puzzles}
\label{sec:ContextRepetition}

As shown in Table~\ref{tab:language_cases}, all three GPT-4o \textbf{One Shot} puzzles, and the Greek OpenAI's o1 \textbf{Few Shot} puzzle do not require any analysis of the puzzle context. Rather, their questions request the repetition of the examples used in the puzzle context. An example of such a puzzle is the Greek OpenAI o1 \textbf{Few Shot} puzzle presented in App.\ref{sec:AppB}.
The questions generated for this puzzle ask the participant to translate into Greek (in Roman script) the following four English phrases: (1) The small woman; (2) The small man; (3)  The child; (4) The small child. 
The solutions for all these questions are presented \textit{verbatim} in the puzzle context.

\subsection{External Knowledge}
\label{sec:InvalidPuzzles}

All three \textbf{Zero Shot} GPT-4o and all three \textbf{One Shot} OpenAI's o1 puzzles  (App.\ref{sec:AppB}) are invalid according to the third criterion listed by Zhurinsky: solving them requires external language knowledge. For example, the GPT-4o \textbf{Zero Shot} Spanish puzzle lists only Spanish adjectives. However, the questions ask for the translations of noun phrases, which require knowledge of Spanish articles and nouns. This situation is similar to the Greek puzzles analyzed by the evaluators (see Table~\ref{tab:iaa_results}). 

\subsection{Valid Puzzles}
\label{sec:EasyValidPuzzles}

Several generated puzzle can be marked as easy. However, 
this outcome is promising as it suggests LLMs' potential to generate valid puzzles. One example of a generated valid puzzles is the Spanish OpenAI's o1 \textbf{Few Shot} puzzle presented in App.\ref{sec:AppB}.
%
The question asks to translate four English sentences into Spanish: (1) The boys are kind; (2) The girl is tall; (3) The (female) teacher is tall; (4) The girls are kind. The solution can be easily deduced from the presented puzzle context.
%

One observation from Table~\ref{tab:language_cases} is that, in most settings, the puzzles generated for all three languages by a particular setting fall into the same group. One possible conclusion is that, at present, LLMs generate puzzles in a language-independent manner. However, for the task of linguistic puzzle generation, language independence is a disadvantage, as the most interesting puzzles are those that capture the unique peculiarities of different languages.

%% file: conclusion.tex
\label{sec:conclusion}


We analyze the performance of LLMs for solving and generating linguistic puzzles. For the novel task of linguistic puzzle generation, LLMs are not yet capable of producing Olympiad-level puzzles. However, we demonstrate that under certain prompt settings, LLMs can generate valid, albeit relatively simple, puzzles. We consider this a promising result for this novel, exciting task. 

Our findings indicate that modern LLMs with reasoning capabilities (e.g., OpenAI’s o1) outperform humans in solving puzzles related to phonology, morphology, compounding, syntax, semantics, and number systems irrespectively of the puzzles difficulty levels. However, for puzzles focused on deciphering writing systems, OpenAI’s o1 surpasses humans only at the two lowest difficulty levels, while humans outperform LLMs at the three higher difficulty levels. This observation is confirmed during the puzzles evaluation process. 


%% file: limitations.tex
\label{sec:limitations}

We identify four main limitations in the puzzle generation procedure described in this paper and believe these limitations are interdependent.

First, the number of puzzles in the \textsc{LingOly} benchmark, on the ILO website, and on national linguistic Olympiad websites is relatively small for an LLM to reliably learn the rules of puzzle generation. A larger dataset is needed to develop a more robust puzzle-generation procedure. The more effective this procedure becomes, the more usable puzzles it can produce.

Second, in this project, we focus solely on generating beginner-level morphology puzzles. As noted in Section~\ref{sec:peopleVsLLM}, an LLM’s performance varies depending on the linguistic topic and difficulty level of the puzzle it is solving. It is possible that puzzle generation is similarly influenced by the linguistic topic. Additionally, our experiments are limited to generating puzzles for only three languages.  

Third, in this work, we evaluate only the \textbf{validity} of the generated puzzles, that is, whether they can be solved using \textbf{only} the provided puzzle context. While we note that the valid generated puzzles tend to be easy, there is no formal evaluation method to assess their difficulty or creativity. We see creativity assessment as a major bottleneck in the task of linguistic puzzle generation. On the one hand, evaluating creativity is inherently subjective. 

Fourth, we believe that the creativity of valid linguistic puzzles can best be judged by expert puzzle creators. However, the number of such experts is very limited. 

%% file: acknowledgments.tex
\label{sec:acknowledgements}

We would like to thank three linguistic competition participants for their help in evaluating the automatically generated puzzles: Jinfan Frank Hu, Anne Huang, and Denys Tereshchenko (listed alphabetically). We are grateful for their valuable input and comments. Frank, Anne, and Denys participated in NACLO (North American Computational Linguistics Open Competition).\footnote{\url{naclo.org}} NACLO is the US-based competition analogous to UKLO. Frank, Anne, and Denys participated in NACLO  multiple times and were among the top performers in 2025. Their extensive experience with linguistics puzzles and their thoughtful feedback on puzzle quality were invaluable to us.

%% file: checklistResNLP.tex
\label{sec:checklistResNLP}

\begin{itemize}
    \item A. For every submission
    \begin{enumerate}
        \item Did you describe the limitations of your work? \textcolor{blue}{[Yes]}
        \item Did you discuss any potential risks of your work? \textcolor{gray}{[N/A]}
    \end{enumerate}
    \item B. Did you use or create scientific artifacts?
    \begin{enumerate}
        \item Did you cite the creators of artifacts you used? \textcolor{blue}{[Yes]} We cite the creators of the LLMs used in Sections \ref{sec:Intro}, \ref{relatedWork}, \ref{sec:dataSet}, \ref{sec:peopleVsLLM}, \ref{sec:theorySection}.
        \item Did you discuss the license or terms for use and / or distribution of any artifacts? \textcolor{blue}{[Yes]}: Sections \ref{sec:Intro}, \ref{relatedWork}.
        \item Did you discuss if your use of existing artifact(s) was consistent with their intended use, provided that it was specified? For the artifacts you create, do you specify intended use and whether that is compatible with the original access conditions (in particular, derivatives of data accessed for research purposes should not be used outside of research contexts)? \textcolor{blue}{[Yes]}: Sections \ref{sec:peopleVsLLM}, \ref{sec:theorySection}.
        \item Did you discuss the steps taken to check whether the data that was collected / used contains any information that names or uniquely identifies individual people or offensive content, and the steps taken to protect / anonymize it? \textcolor{gray}{[N/A]}
        \item Did you provide documentation of the artifacts, e.g., coverage of domains, languages, and linguistic phenomena, demographic groups represented, etc.? \textcolor{blue}{[Yes]}: Sections \ref{sec:dataSet},  \ref{sec:peopleVsLLM}, \ref{sec:theorySection}.
        \item Did you report relevant statistics like the number of examples, details of train / test / dev splits, etc. for the data that you used / created? \textcolor{blue}{[Yes]} We report the relevant statistics in Section \ref{sec:dataSet}, \ref{sec:peopleVsLLM}, \ref{sec:theorySection}.
    \end{enumerate}
    \item C. Did you run computational experiments?
    \begin{enumerate}
        \item Did you report the number of parameters in the models used, the total computational budget (e.g., GPU hours), and computing infrastructure used? \textcolor{gray}{[N/A]}
        \item Did you discuss the experimental setup, including hyperparameter search and best-found hyperparameter values? \textcolor{blue}{[Yes]}: Sections \ref{sec:peopleVsLLM}, \ref{sec:theorySection}.
        \item Did you report descriptive statistics about your results (e.g., error bars around results, summary statistics from sets of experiments), and is it transparent whether you are reporting the max, mean, etc. or just a single run? \textcolor{blue}{[Yes]}: Sections \ref{sec:dataSet}, \ref{sec:peopleVsLLM}, \ref{sec:theorySection}.
        \item If you used existing packages (e.g., for preprocessing, for normalization, or for evaluation, such as NLTK, Spacy, ROUGE, etc.), did you report the implementation, model, and parameter settings used? \textcolor{red}{[No]}
    \end{enumerate}
    \item D. Did you use human annotators (e.g., crowdworkers) or research with human participants?
    \begin{enumerate}
        \item Did you report the full text of instructions given to participants, including e.g., screenshots, disclaimers of any risks to participants or annotators, etc.? \textcolor{blue}{[Yes]}: Section \ref{sec:theorySection}.
        \item Did you report information about how you recruited (e.g., crowdsourcing platform, students) and paid participants, and discuss if such payment is adequate given the participants’ demographic (e.g., country of residence)? \textcolor{blue}{[Yes]}: Section \ref{sec:theorySection}.
        \item Did you discuss whether and how consent was obtained from people whose data you’re using/curating? \textcolor{blue}{[Yes]}: Section \ref{sec:theorySection}.
        \item Was the data collection protocol approved (or determined exempt) by an ethics review board? \textcolor{gray}{[N/A]} Our experiment falls under one of the exempt categories as per human subject research handbook.
        \item Did you report the basic demographic and geographic characteristics of the annotator population that is the source of the data? \textcolor{blue}{[Yes]} We mention this in Section \ref{sec:theorySection}.
    \end{enumerate}
    \item E. Did you use AI assistants (e.g., ChatGPT, Copilot) in your research, coding, or writing?
    \begin{enumerate}
        \item Did you include information about your use of AI assistants? \textcolor{blue}{[Yes]} LLMs are used in the experiments described in the paper.
    \end{enumerate}
\end{itemize}

%% file: appendix_a.tex
\label{sec:AppA}

\begin{figure}[htbp]
\centering
 \textbf{Xhosa puzzle:} UKLO, 2024
 \includegraphics[width=\columnwidth]{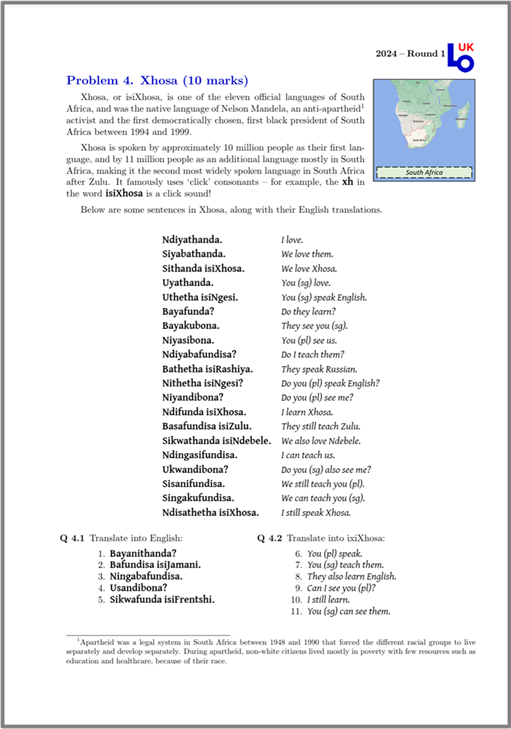}
 \caption{The Xhosa puzzle was used in UKLO in 2024. This puzzle has two difficulty scores: its score for the Foundation participants is $58\%$ and its score for the Intermediate participants $81\%$; its linguistic topic is morphology; its type is Rosetta; its language family is Atlantic–Congo, Bantu; its Author is Babette Verhoeven. \\ \url{https://www.uklo.org/wp-content/uploads/2024/04/2024_R1_4-Xhosa.pdf}}  
  \label{fig:Xhosa}
\end{figure}

\vfill\eject

\begin{figure}[htbp]
\vspace{47pt}
\centering
 \textbf{Waama puzzle:} UKLO, 2021
  \includegraphics[width=\columnwidth]{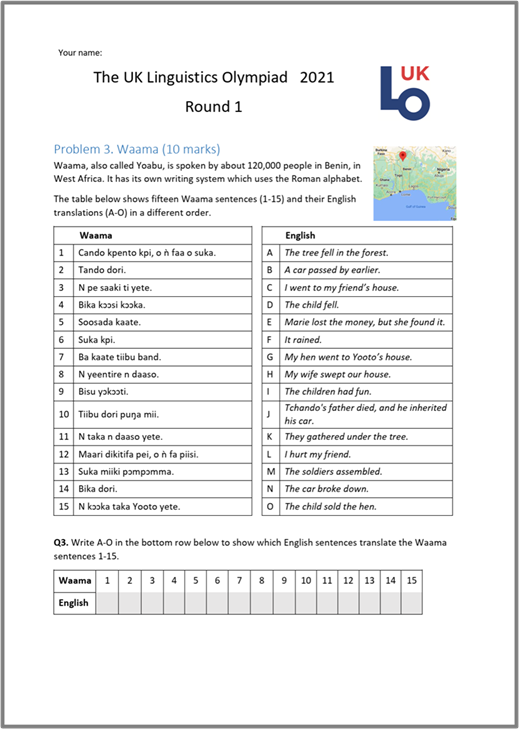}
  \caption{The Waama puzzle was used in UKLO in 2021. This puzzle has two difficulty scores: its score for the Breakthrough participants is $42\%$ and its score for the Foundation participants $54\%$; its linguistic topic is Syntax; its type is Match-up; its language family is Atlantic–Congo, Gur; its Author is Aleka Blackwell. \\ 
  \url{https://www.uklo.org/wp-content/uploads/2022/05/2021_3-Waama.pdf}}
  \label{fig:Waama}
\end{figure}

\newpage

\begin{figure}[htbp]
\centering
    \textbf{Warlpiri puzzle:} UKLO, 2024
  \includegraphics[width=\columnwidth]{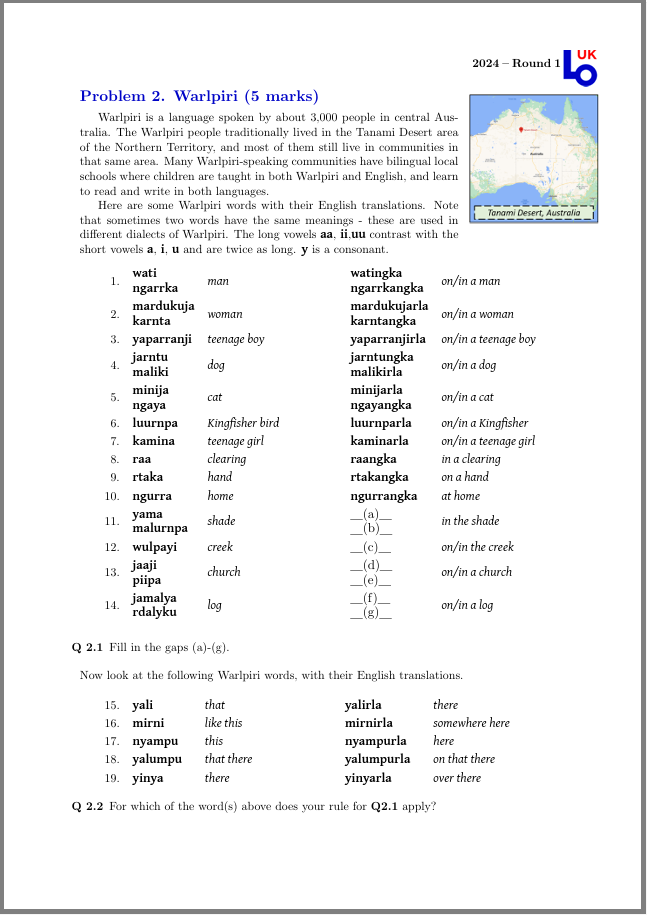}
  \caption{The Warlpiri puzzle was used in UKLO in 2024. This puzzle has two difficulty scores: its score for the Breakthrough participants is $41\%$ and its score for the Foundation participants $45\%$; its linguistic topic is a combination of morphology and phonology; its type is Pattern; its language family is Pama-Nyungan; its Author is Mary Laughren.  \\ \url{https://www.uklo.org/wp-content/uploads/2024/04/2024_R1_2-Warlpiri.pdf}}
  \label{fig:Warlpiri}
\end{figure}

\vfill\eject

\begin{figure}[htbp]
\centering
    \textbf{Wik-Mungkan puzzle:} UKLO, 2022
  \includegraphics[width=\columnwidth]{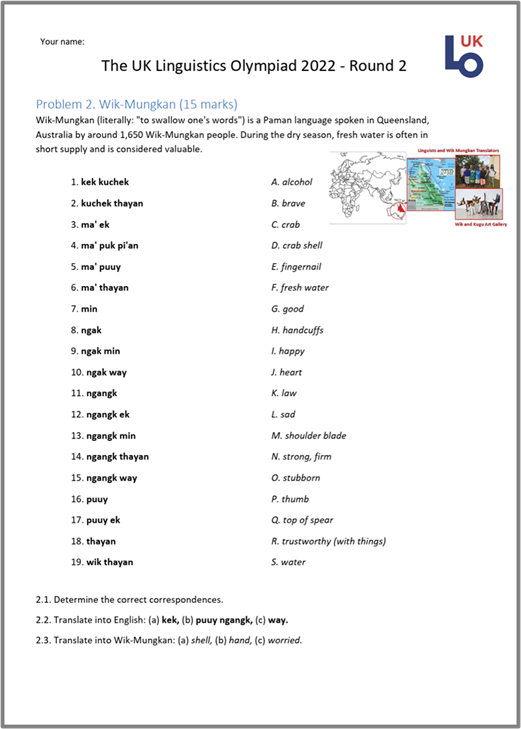}
  \caption{The Wik-Mungkan puzzle was used in Round 2 of UKLO in 2022. Its score for participants is $28\%$; its linguistic topic is Compounding; its type is Match-up; its language family is Pama-Nyungan; its Author is Ryan Chi. \\ \url{https://www.uklo.org/wp-content/uploads/2022/05/2022_R2_2_Wik-Mungkan.pdf}}
  \label{fig:WikMungkan}
\end{figure}


\newpage

\begin{figure}[htbp]
\centering
    \textbf{Ditema puzzle:} UKLO, 2019
  \includegraphics[width=\columnwidth]{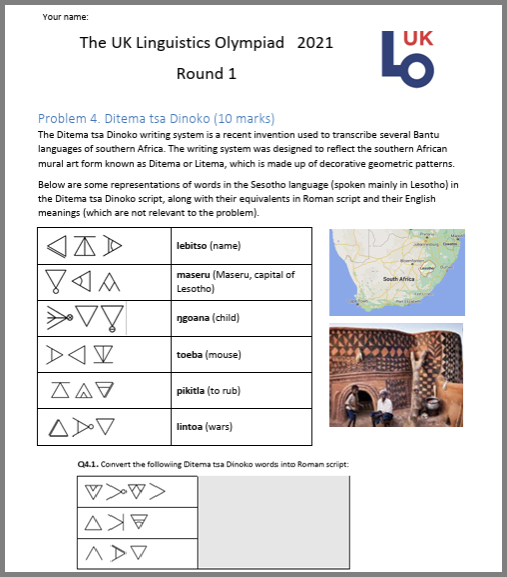}
  \caption{Ditema puzzle was used in UKLO in 2019. This puzzle has two difficulty scores: its score for the Foundation participants is $28\%$, its score for the Intermediate participants is $51\%$; its linguistic topic is writing system; its type is Rosetta; its language family is Atlantic–Congo, Bantu; its author is Michael Salter.\\
  \url{https://www.uklo.org/wp-content/uploads/2022/05/2021_4-Ditema.pdf}}
  \label{fig:Ditema}
\end{figure}


\vfill\eject

\begin{figure}[htbp]
\centering
    \textbf{Georgian puzzle:} UKLO, 2015
  \includegraphics[width=\columnwidth]{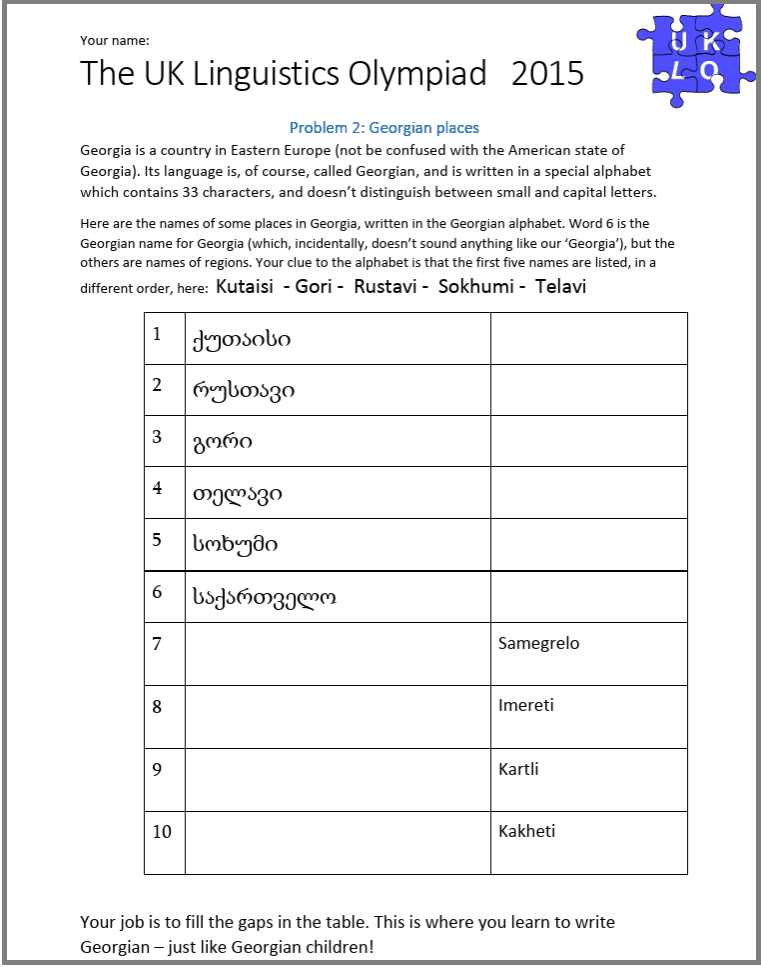}
  \caption{The Georgian puzzle was used in UKLO in 2015. This puzzle has two difficulty scores: its score for the Breakthrough participants is $71\%$, its score for the Foundation participants is $79\%$; its linguistic topic is writing system; its type is Match-up; its language family is Kartvelian; its Author is Daniel Rucki. \\ \url{https://www.uklo.org/wp-content/uploads/2022/05/2015_2.-Georgian.pdf}}
  \label{fig:Georgian}
\end{figure}

\newpage

\begin{figure}[htbp]
\centering
    \textbf{Maonan puzzle:} UKLO, 2024
  \includegraphics[width=\columnwidth]{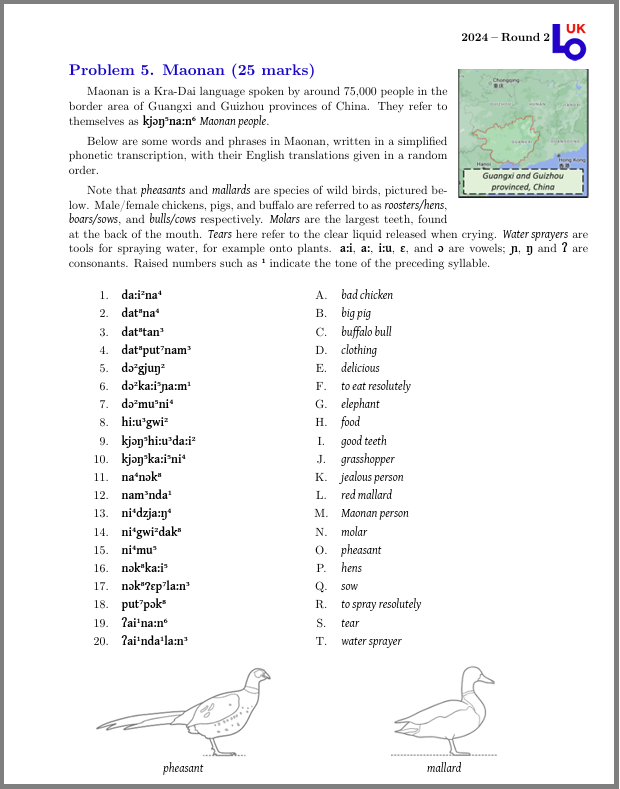}
  \caption{The Maonan puzzle was used in Round 2 of UKLO in 2024. Its score for participants is $5\%$; its linguistic topic is a combination of Semantics and Compounding; its type is Match-up; its language family is Kra-Dai; its Author is Daniel Titmas. \\ \url{https://www.uklo.org/wp-content/uploads/2024/03/2024_R2_5-Maonan.pdf}}
  \label{fig:Maonan}
\end{figure}

\vfill\eject

\begin{figure}[htbp]
\centering
    \textbf{Ngkolmpu puzzle:} UKLO, 2021
  \includegraphics[width=\columnwidth]{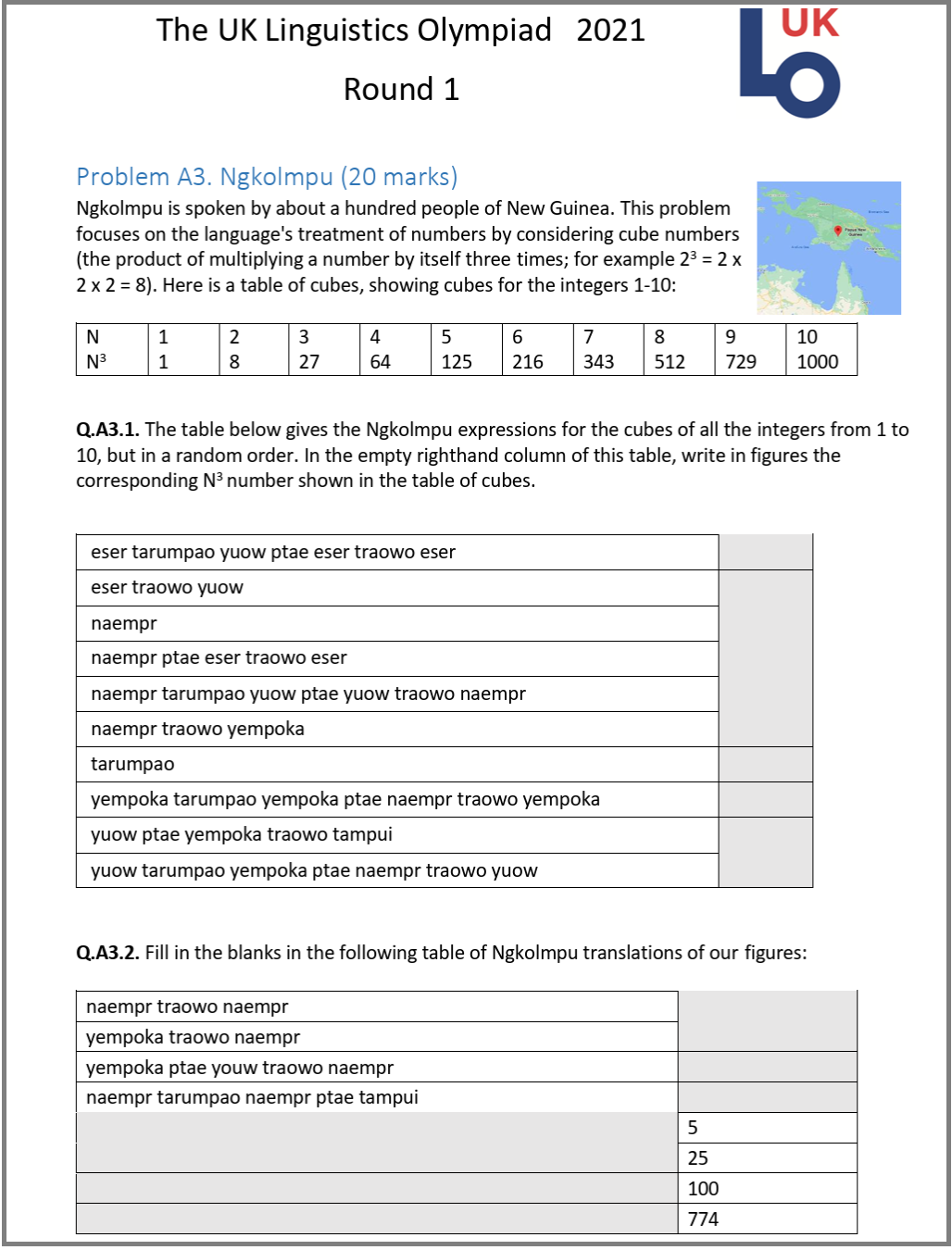}
  \caption{The Ngkolmpu  puzzle was used in UKLO in 2021. Its difficulty level is Advanced. Its score for participants is $35\%$; its linguistic topic is numeric system; its type is Match-up; its language family is Yam; its Author isSimi Hellsten. \\ \url{https://www.uklo.org/wp-content/uploads/2022/05/2021_A3-Ngkolmpu.pdf}}
  \label{fig:Ngkolmpu}
\end{figure}

\newpage

\begin{figure}[htbp]
\centering
    \textbf{Mazateco puzzle:} UKLO, 2022
  \includegraphics[width=\columnwidth]{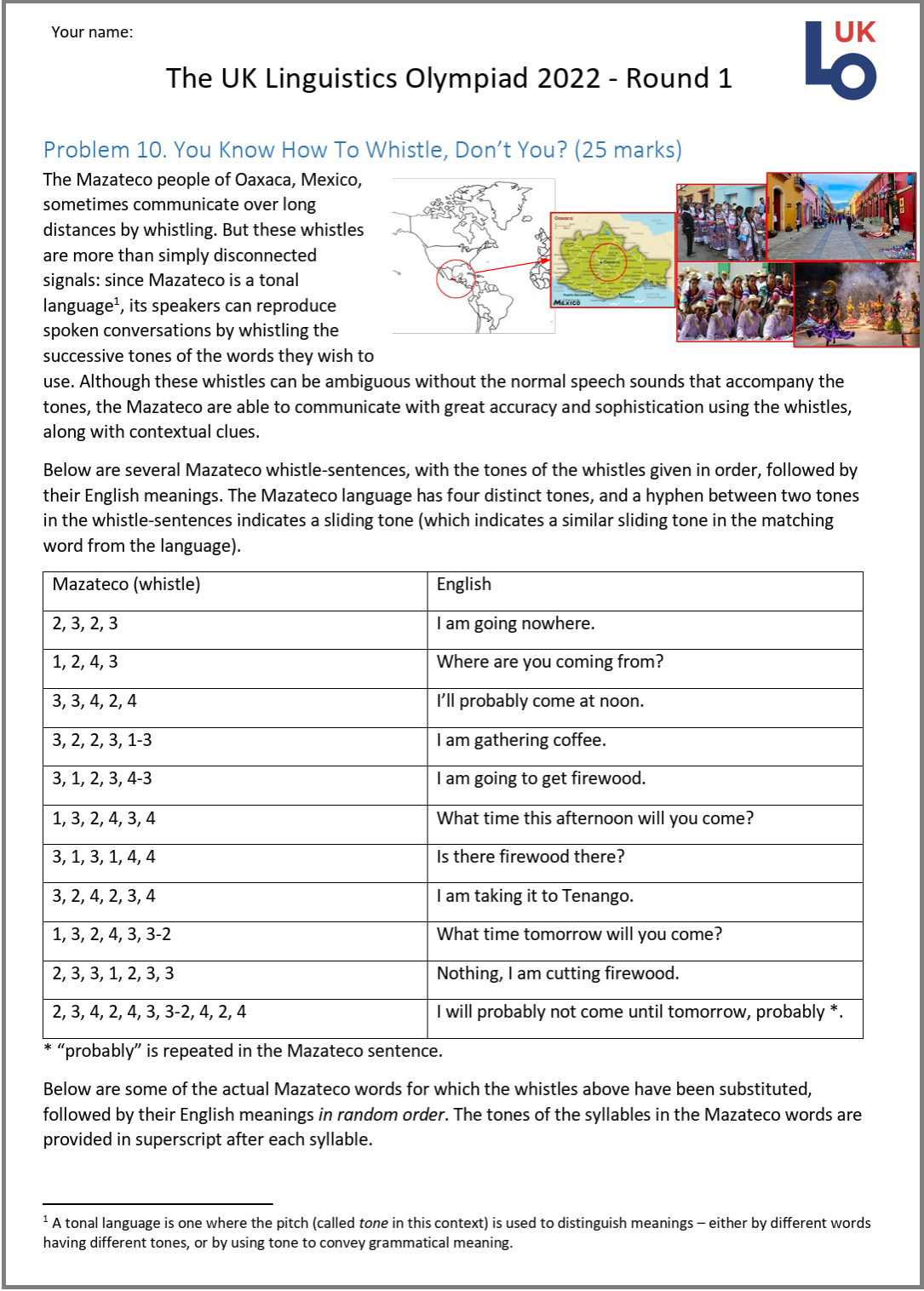}
  \caption{The Mazateco puzzle was used in UKLO in 2022. Its difficulty level is Advanced. Its score for participants is $37\%$; its linguistic topic is Syntax; its type is a combination Match-up and Rosetta; its language family is Otomanguean; its Author is Michael Salter. \\ \url{https://www.uklo.org/wp-content/uploads/2022/05/10_Adv_UKLO-2022-Mazateco_You-Know-How-To-Whistle-Dont-You_Complete-Script.pdf}}
  \label{fig:Mazateco}
\end{figure}

\vfill\eject

\begin{figure}[htbp]
\centering
    \textbf{Maltese puzzle:} UKLO, 2022
  \includegraphics[width=\columnwidth]{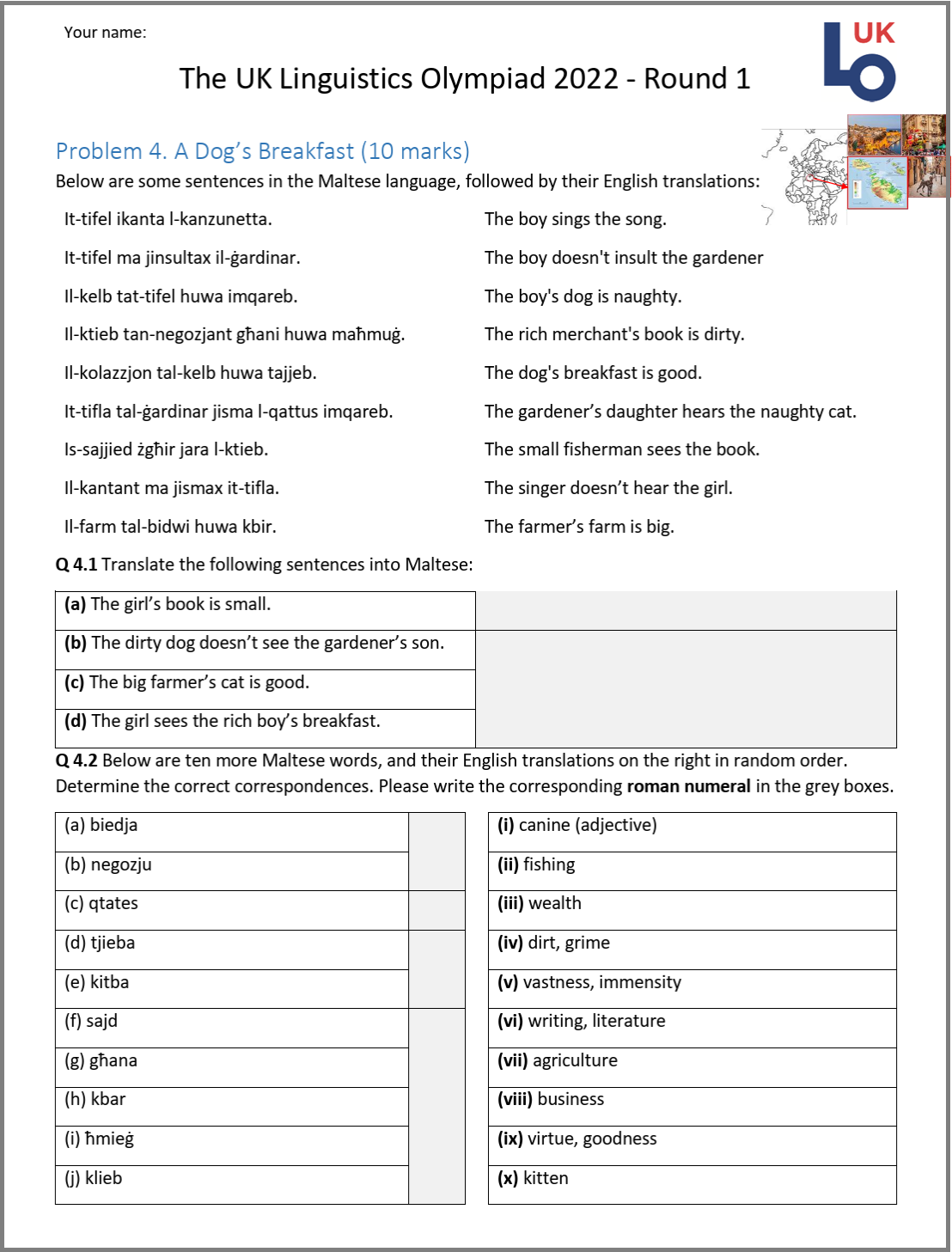}
  \caption{The Mazateco puzzle was used in UKLO in 2022. This puzzle has two difficulty scores: its score for the Foundation participants is $58\%$, its score for the Intermediate participants is $79\%$; ; its linguistic topic is a combination Phonology, Syntax, and Morphology; its type is a combination Match-up and Rosetta; its language family is Afro-Asiatic, Semitic; its Author is Michael Salter. \\ 
  \url{https://www.uklo.org/wp-content/uploads/2022/06/4_UKLO-2022-Maltese_A-Dogs-Breakfast_-Complete-Script.pdf}}
  \label{fig:Maltese}
\end{figure}

\newpage

\begin{figure}[htbp]
\centering
    \textbf{Lithuanian puzzle (preamble and context)} 
  \includegraphics[width=\columnwidth]{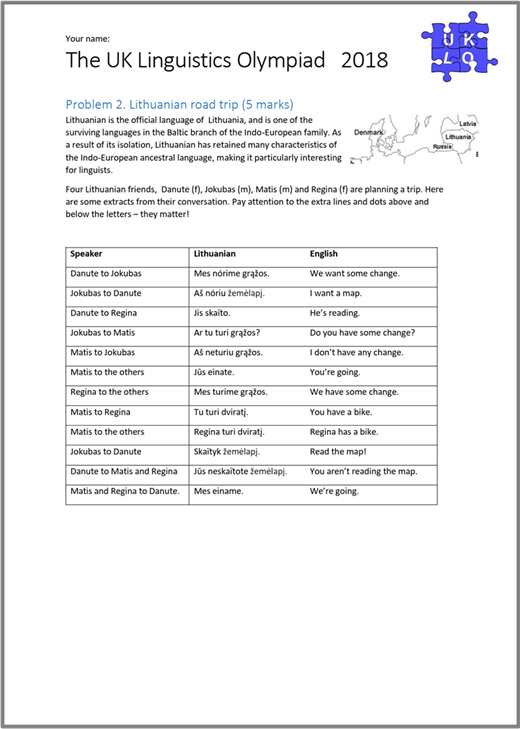}
  \caption{The Lithuanian puzzle was used in UKLO in 2018.This puzzle has two difficulty scores: its score for the Breakthrough participants is $40\%$, its score for the Foundation participants is $53\%$; its linguistic topic is a combination of morphology and syntax; its type is Rosetta; its language family is Indo-European, Balto-Slavic; its Author is Babette Verhoeven. \\ 
  \url{https://www.uklo.org/wp-content/uploads/2022/05/2018_2-Lithuanian.pdf}}
  \label{fig:LithuanianContext}
\end{figure}

\vfill\eject

\begin{figure}[htbp]
\centering
    \textbf{Lithuanian puzzle (questions):} UKLO, 2018
  \includegraphics[width=\columnwidth]{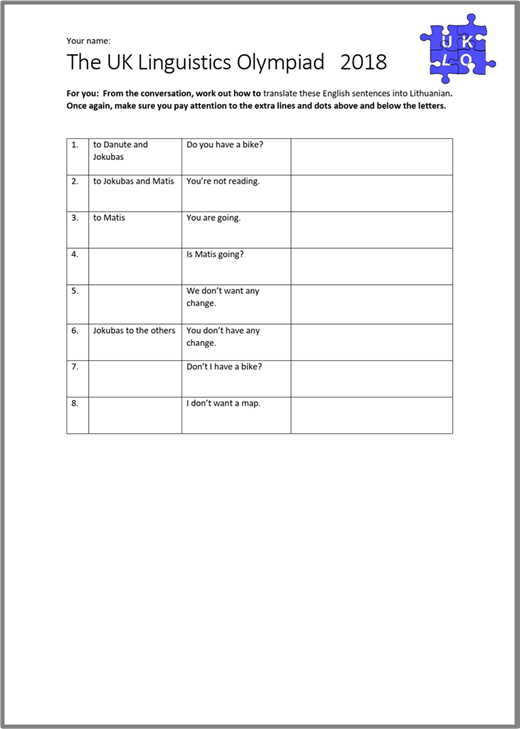}
  \caption{The Lithuanian puzzle was used in UKLO in 2018.This puzzle has two difficulty scores: its score for the Breakthrough participants is $40\%$, its score for the Foundation participants is $53\%$; its linguistic topic is a combination of morphology and syntax; its type is Rosetta; its language family is Indo-European, Balto-Slavic; its Author is Babette Verhoeven. \\ 
  \url{https://www.uklo.org/wp-content/uploads/2022/05/2018_2-Lithuanian.pdf}}
  \label{fig:LithuanianSolution}
\end{figure}

\newpage

\begin{figure}[htbp]
\centering
    \textbf{Swedish puzzle:} UKLO, 2022
  \includegraphics[width=\columnwidth]{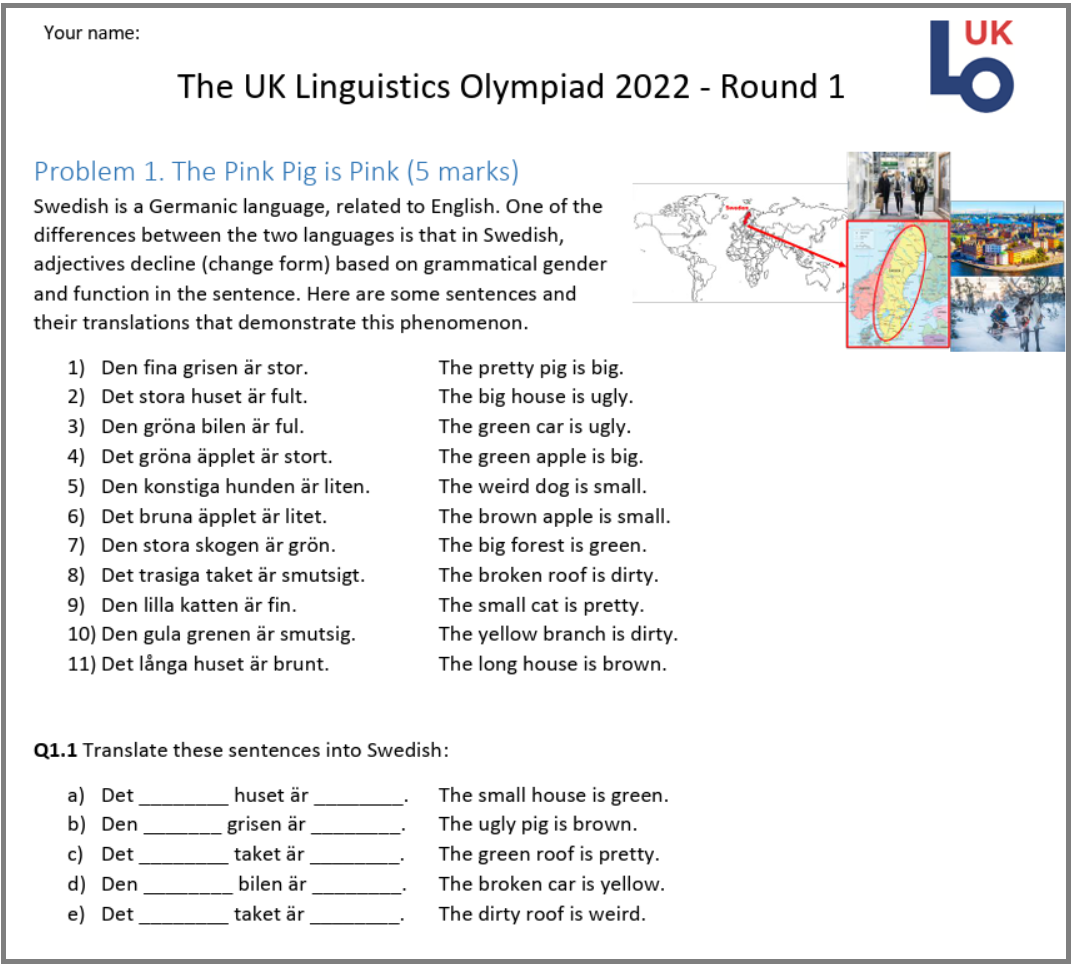}
  \caption{The Swedish puzzle was used in UKLO in 2022. Its difficulty level is Breakthrough. Its score for participants is $38\%$; its linguistic topic is Morphology; its type is Rosetta; its language family is Indo-European, Germanic; its Author is David Hellsten. \\ \url{https://www.uklo.org/wp-content/uploads/2022/05/1_UKLO-2022-Swedish_The-Pink-Pig-is-Pink_-Complete-Script.pdf}}
  \label{fig:Swedish}
\end{figure}

\vfill\eject

\begin{figure}[htbp]
\centering
    \textbf{Kabyle puzzle:} UKLO, 2022
  \includegraphics[width=\columnwidth]{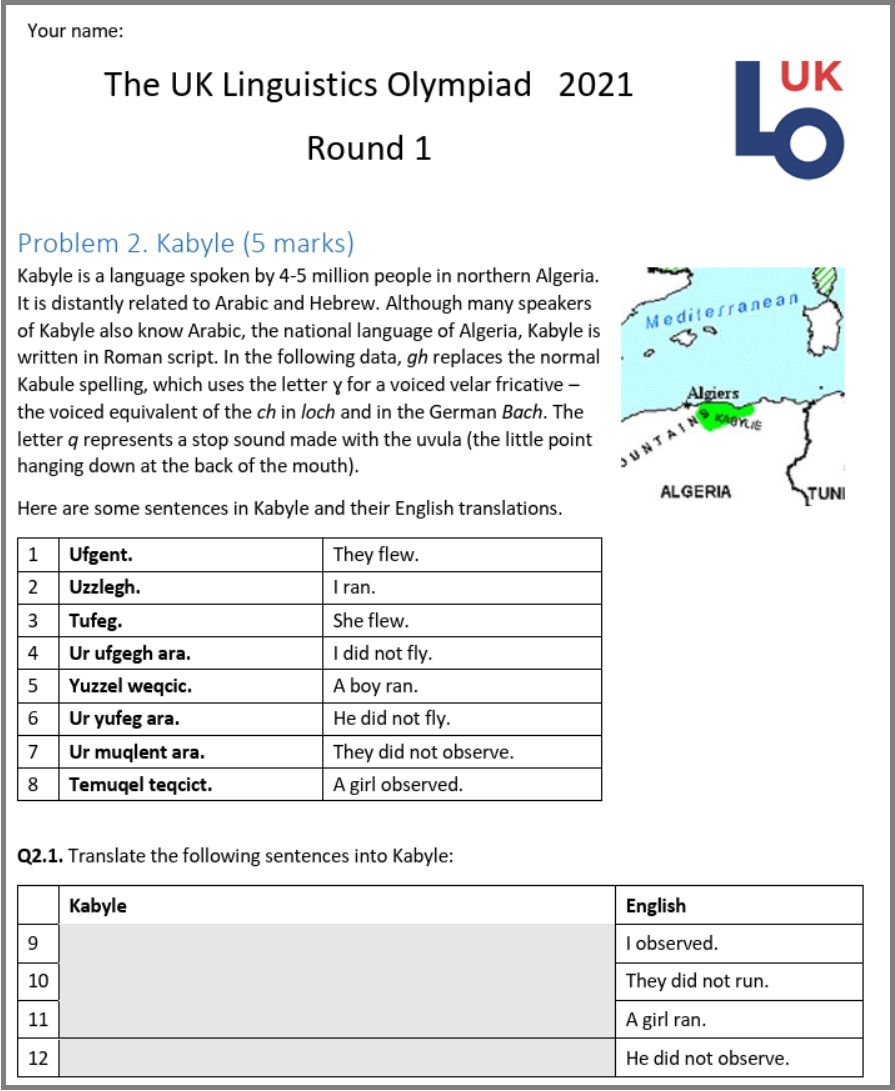}
  \caption{The Kabyle puzzle was used in UKLO in 2021. This puzzle has two difficulty scores: its score for the Breakthrough participants is $44\%$, its score for the Foundation participants is $51\%$; ; its linguistic topic is a combination Syntax and Morphology; its type is Rosetta; its language family is Afro-Asiatic, Semitic; its Authors are Kazune Sato, Simi Hellsten. \\ 
  \url{https://www.uklo.org/wp-content/uploads/2022/05/2021_2-Kabyle.pdf}}
  \label{fig:Kabyle}
\end{figure}

\newpage

\textbf{Greek puzzle}, parallel to the {Georgian puzzle}: UKLO, 2015 example

Greece is a country in Southern Europe. Its language is, of course, called Greek, and is written in a special alphabet which contains 24 characters. Greek distinguishes between small and capital letters; however, in this problem, all the letters are capital letters. Here are the names of some locations written in the Greek alphabet.

Word 6 is the Greek name for Greece (which, incidentally, doesn’t sound anything like our ‘Greece’), but the others are names of regions. Your clue to the alphabet is that the first five names are listed, in a different order, here: Ikaria, Arta, Kalamata, Pisidia, Kea.

\begin{table}[H]
\begin{tabular}{ c | l l }
\hline
& Greek  & English  \\ 
 \hline
1&\foreignlanguage{greek}{ΑΡΤΑ} & Arta          \\
2&\foreignlanguage{greek}{ΙΚΑΡΙΑ} & Ikaria       \\
3&\foreignlanguage{greek}{ΚΑΛΑΜΑΤΑ} & Kalamata   \\
4&\foreignlanguage{greek}{ΚΕΑ} & Kea       \\
5&\foreignlanguage{greek}{ΠΙΣΙΔΙΑ} & Pisidia       \\
7&\foreignlanguage{greek}{ΕΛΛΑΔΑ} &    \\
 \hline
\end{tabular}
\caption{Greek Example.}
\label{tab:GreekPuzzle}
\end{table}

Your job is to fill i the gap in the table. 

\newpage

%% file: appendix_b.tex
\label{sec:AppB}

\subsection{Puzzles Generated by GPT-4o}
\label{sec:4o}




\subsubsection{GPT-4o, Zero-shot, Greek}
\begin{figure}[htbp]
    \centering
    \includegraphics[scale=1]{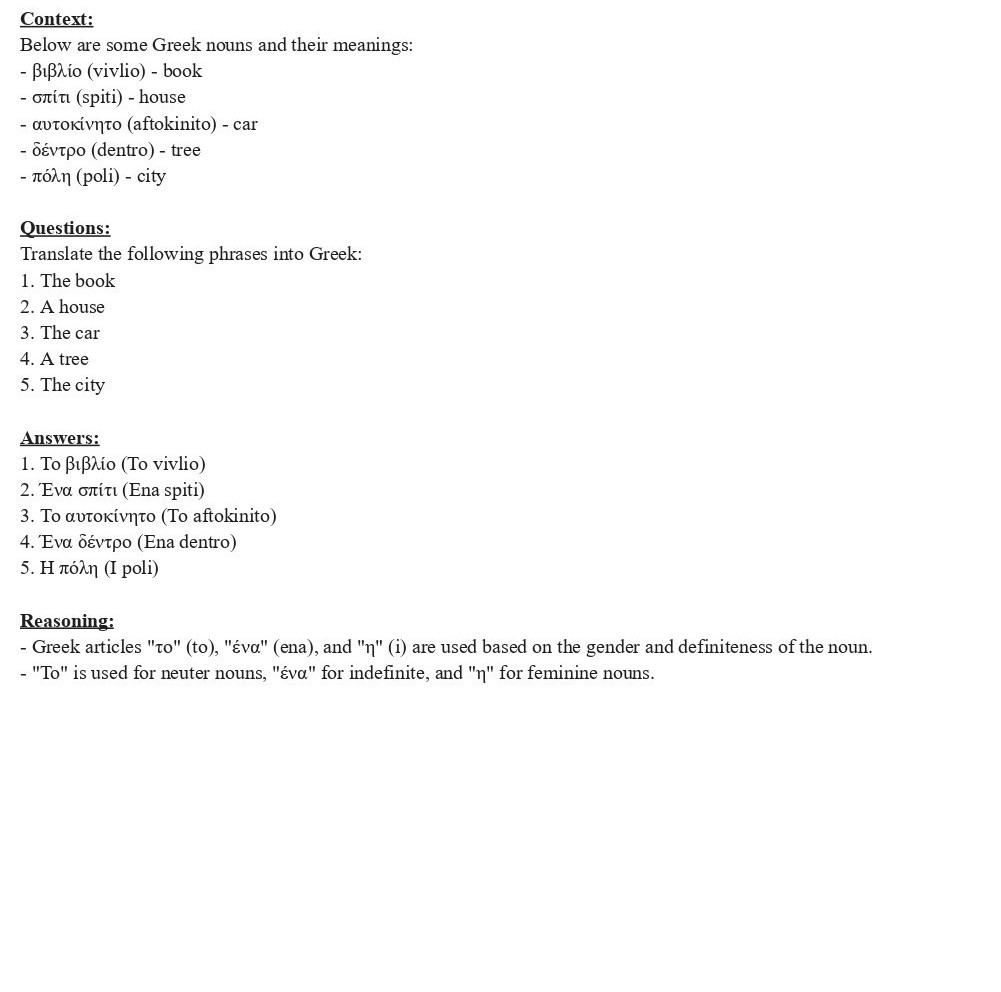}
\end{figure}
\clearpage

\subsubsection{GPT-4o, Zero-shot, Gujarati}
\begin{figure}[htbp]
    \centering
    \includegraphics[scale=1]{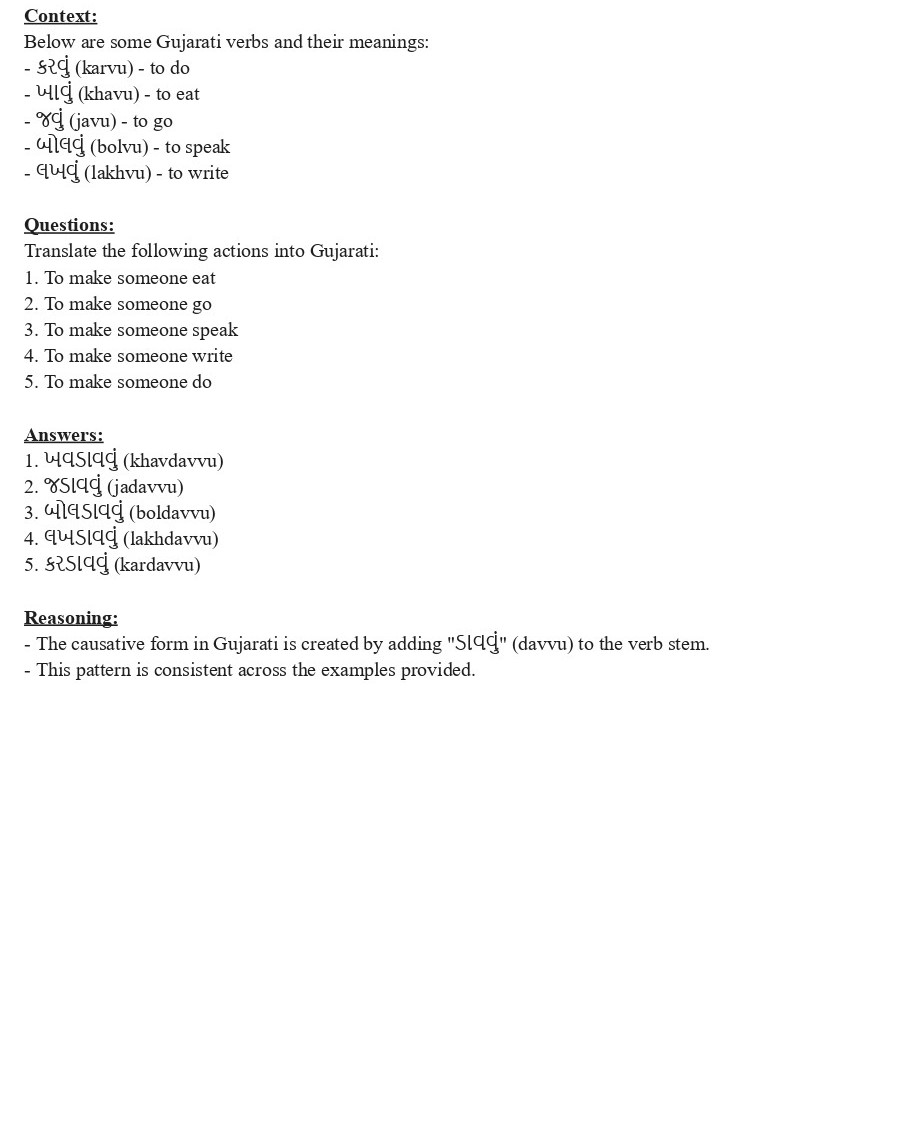}
\end{figure}
\clearpage

\subsubsection{GPT-4o, Zero-shot, Spanish}
\begin{figure}[htbp]
    \centering
    \includegraphics[scale=1]{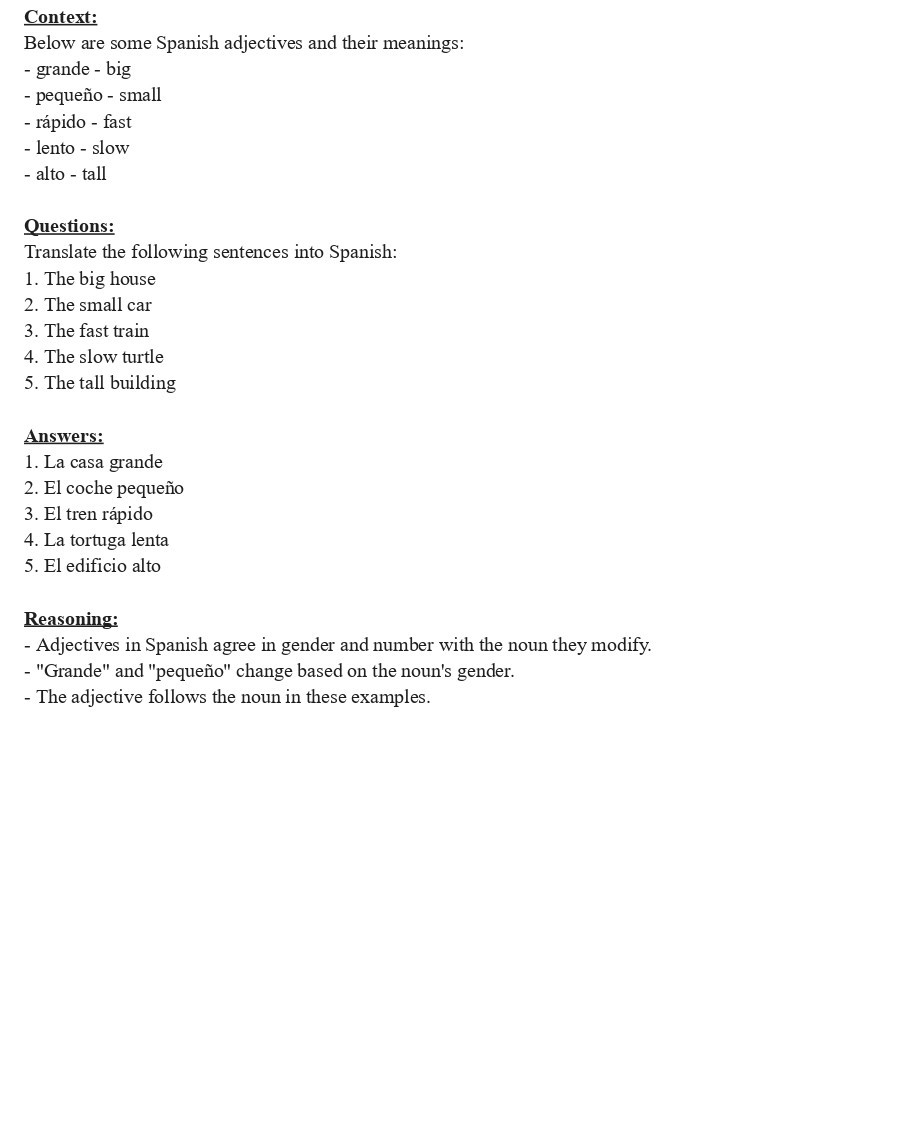}
\end{figure}
\clearpage

\subsubsection{GPT-4o, One-shot, Greek}
\begin{figure}[htbp]
    \centering
    \includegraphics[scale=0.825]{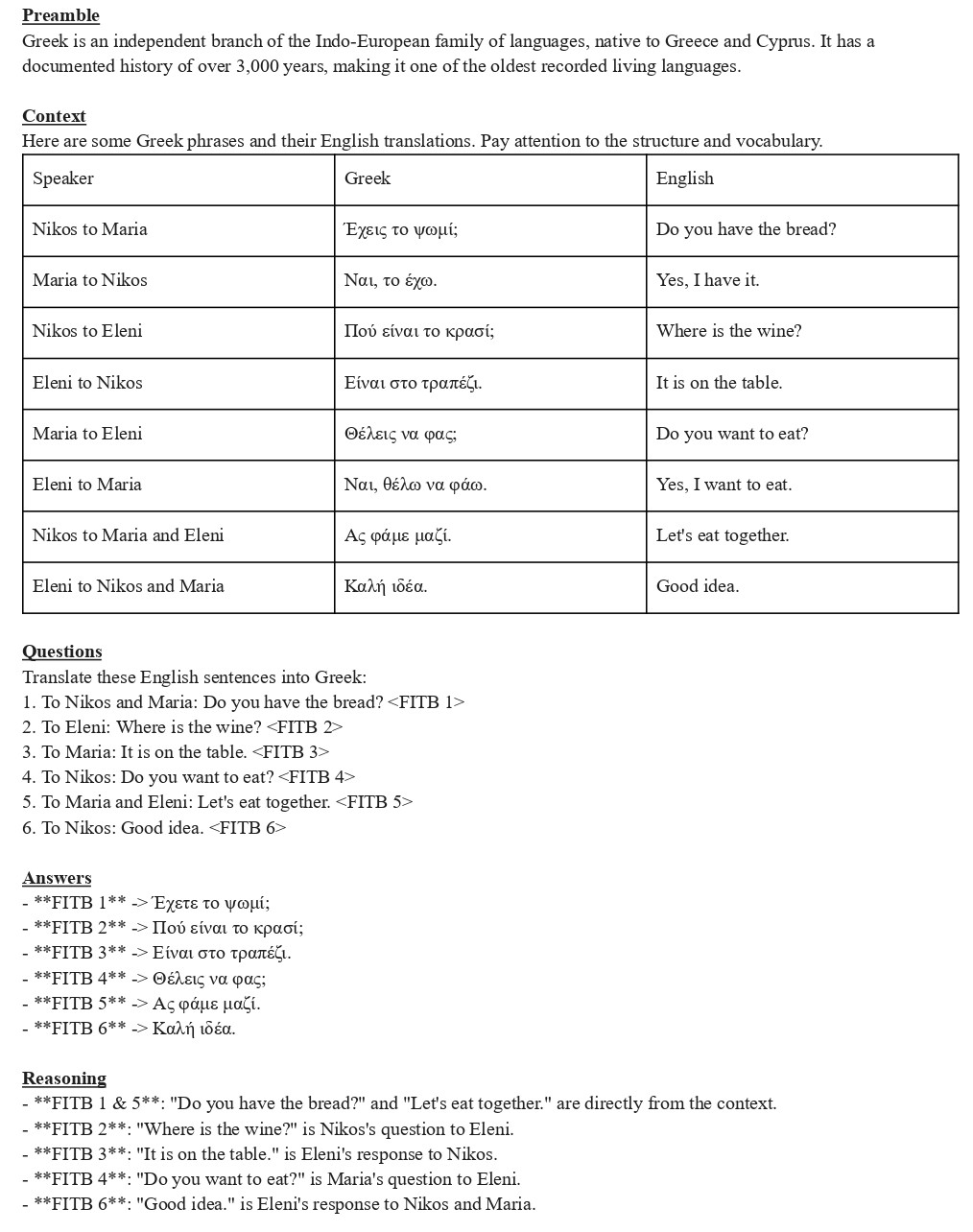}
\end{figure}
\clearpage

\subsubsection{GPT-4o, One-shot, Gujarati}
\begin{figure}[htbp]
    \centering
    \includegraphics[width=\textwidth, height=\textheight, keepaspectratio]{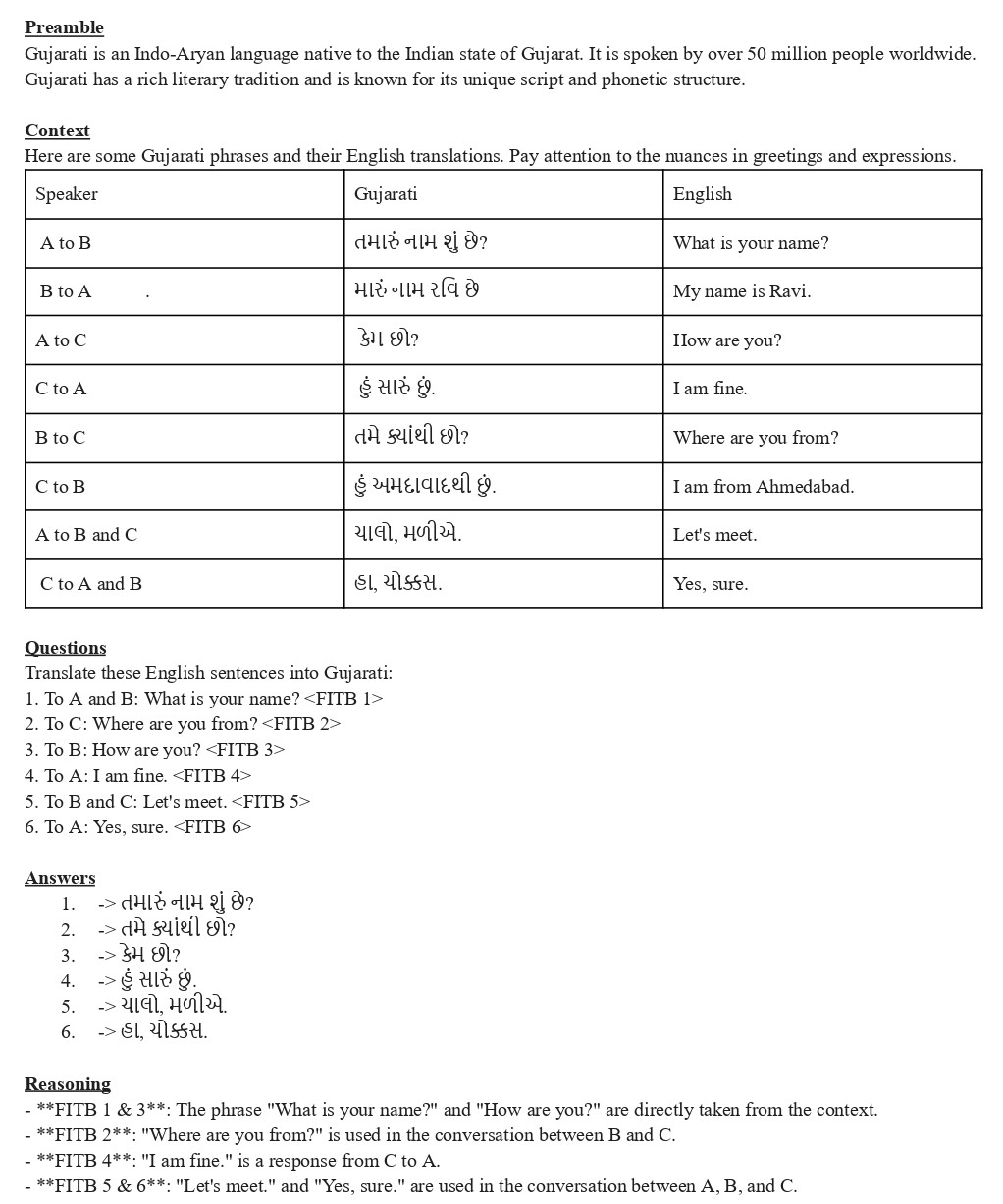}
\end{figure}
\clearpage

\subsubsection{GPT-4o, One-shot, Spanish}
\begin{figure}[htbp]
    \centering
    \includegraphics[scale=0.885]{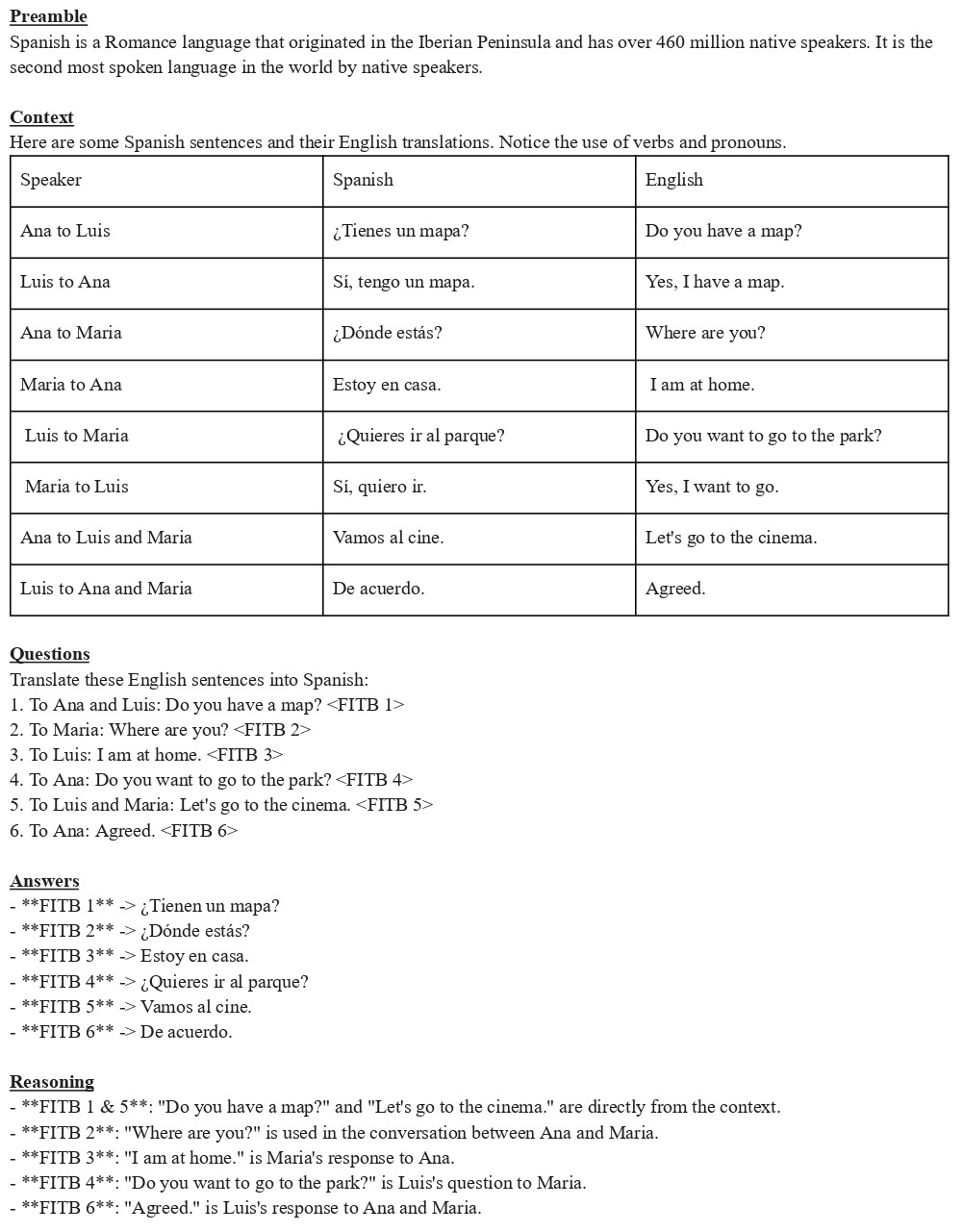}
\end{figure}
\clearpage

\subsubsection{GPT-4o, Few-shot, Greek}
\begin{figure}[htbp]
    \centering
    \includegraphics[scale=0.885]{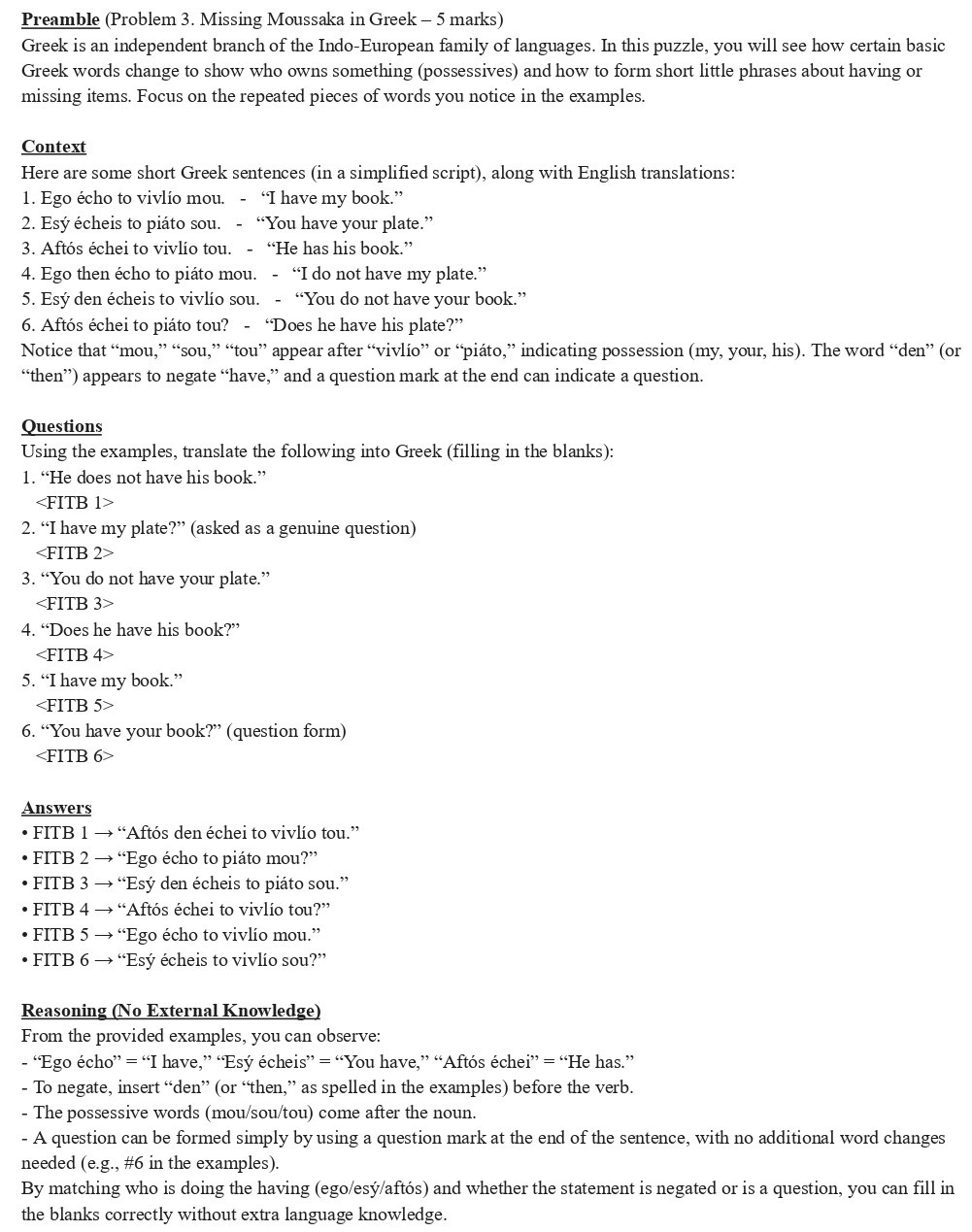}
\end{figure}
\clearpage

\subsubsection{GPT-4o, Few-shot, Gujarati}
\begin{figure}[htbp]
    \centering
    \includegraphics[scale=0.9]{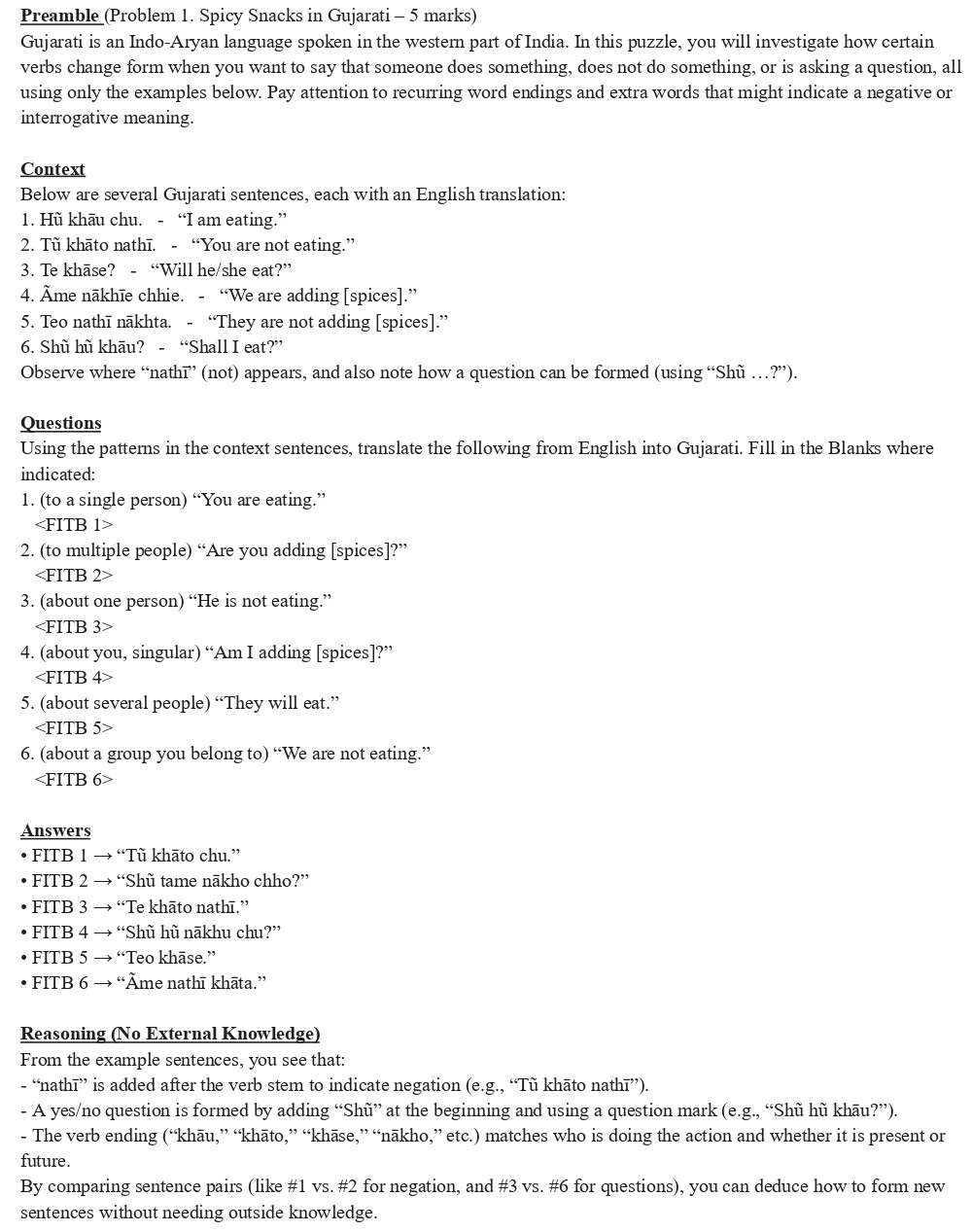}
\end{figure}
\clearpage

\subsubsection{GPT-4o, Few-shot, Spanish}
\begin{figure}[htbp]
    \centering
    \includegraphics[scale=0.915]{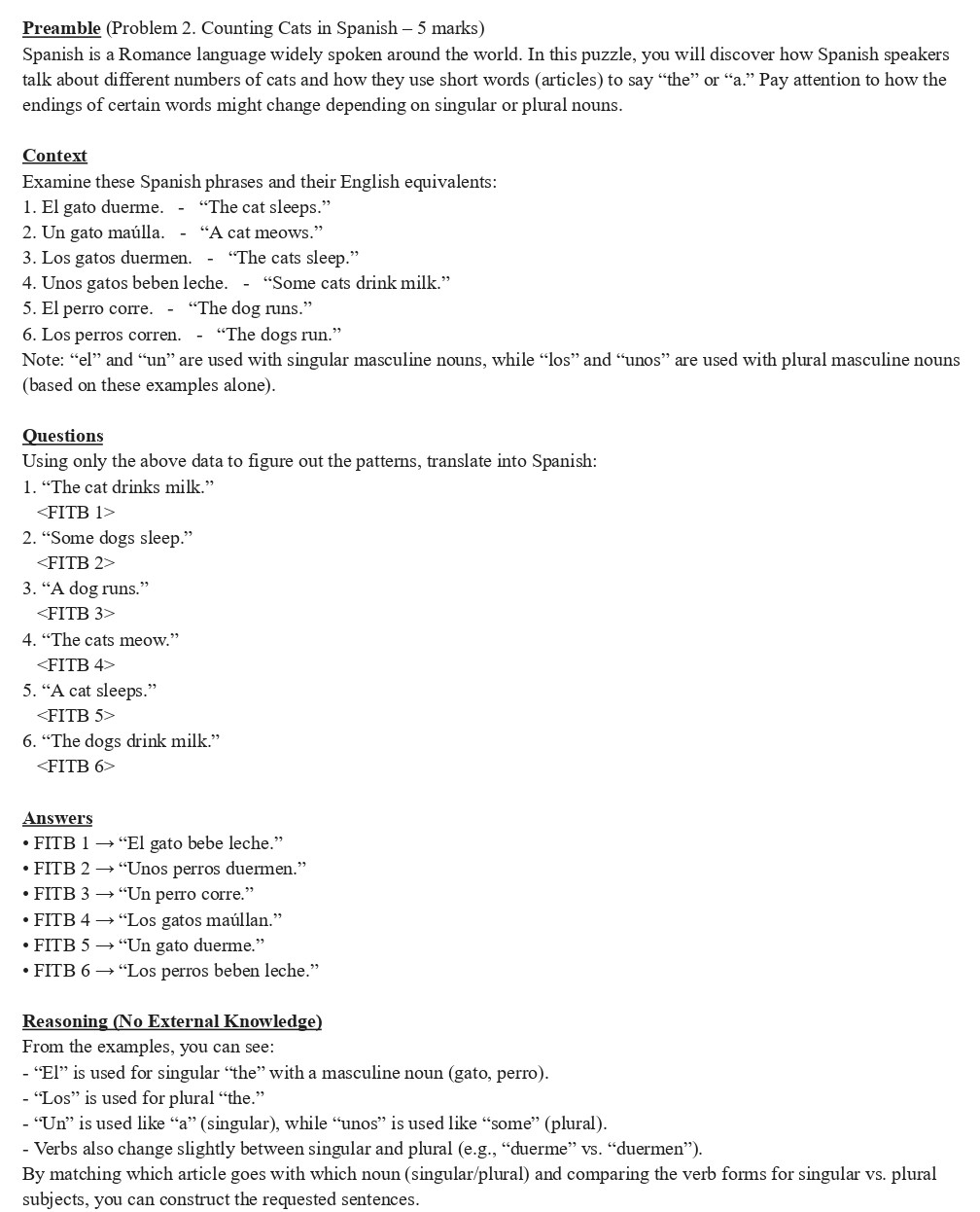}
\end{figure}

\clearpage

\subsection{Puzzles Generated by OpenAI's o1}
\label{sec:o1}
\subsubsection{OpenAI's o1, Zero-shot, Greek}
\begin{figure}[htbp]
    \centering
    \includegraphics[scale=0.8395]{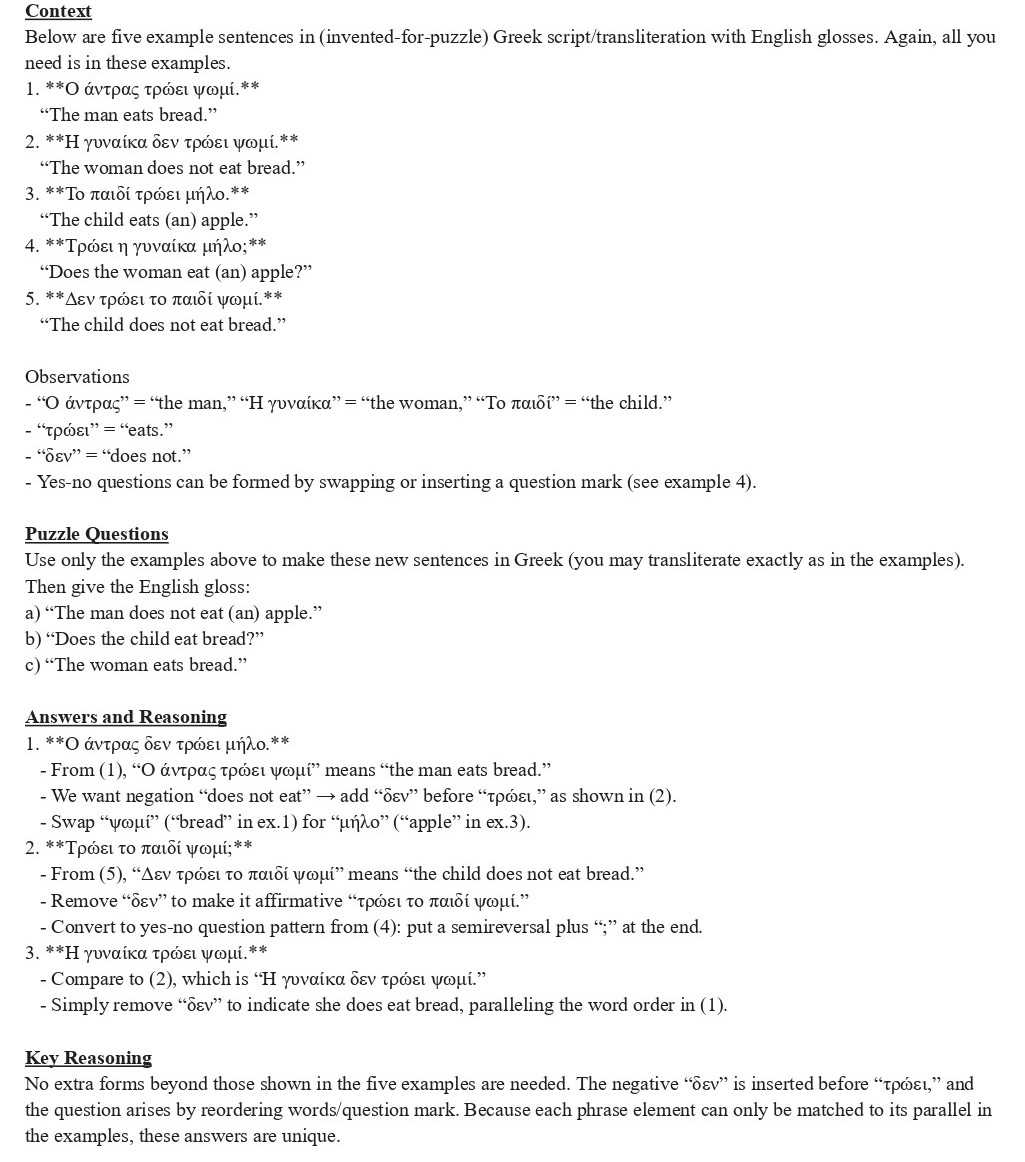}
\end{figure}

\clearpage
\subsubsection{OpenAI's o1, Zero-shot, Gujarati}
\begin{figure}[htbp]
    \centering
    \includegraphics[scale=0.885]{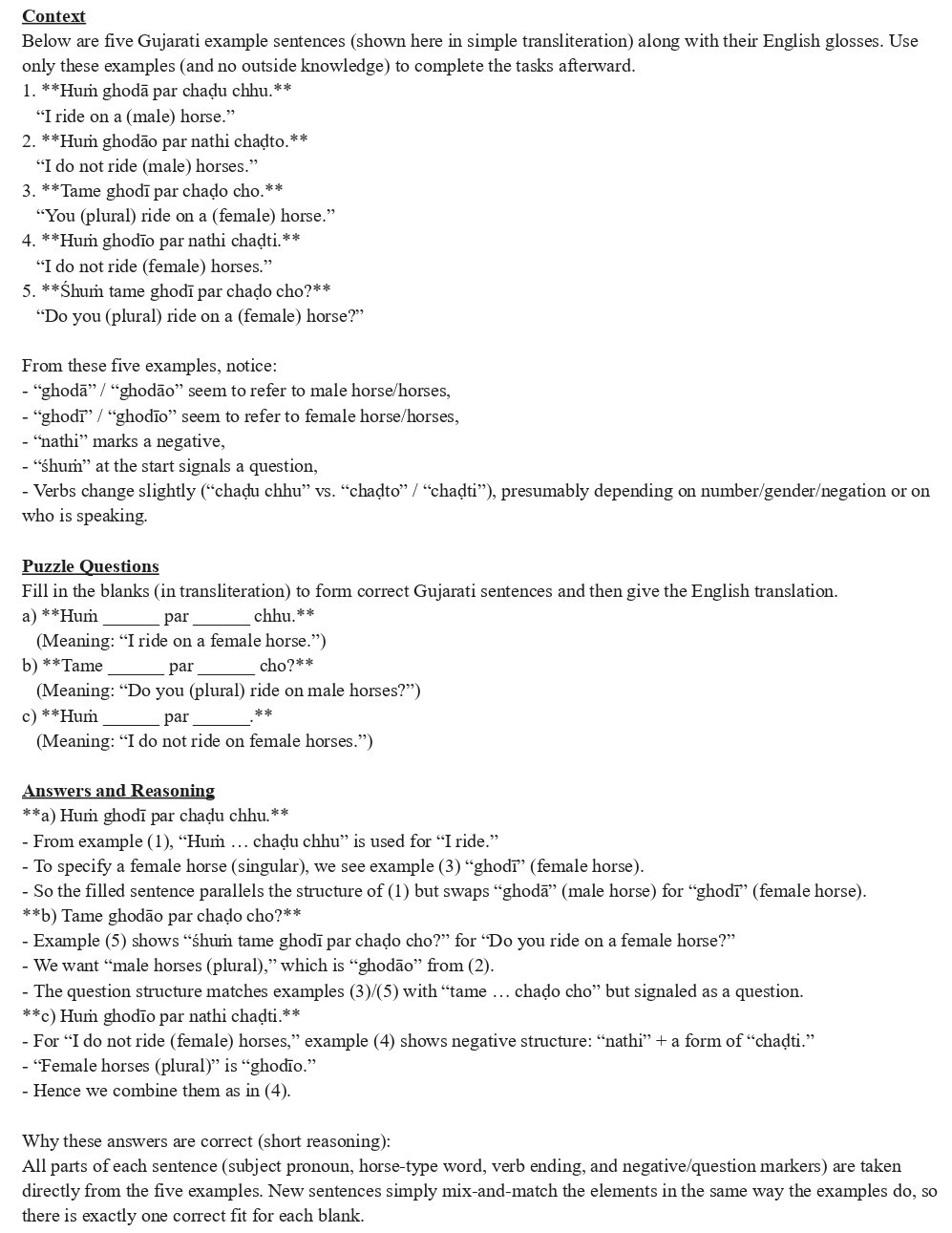}
\end{figure}
\clearpage

\subsubsection{OpenAI's o1, Zero-shot, Spanish}
\begin{figure}[htbp]
    \centering
    \includegraphics[width=\textwidth, height=\textheight, keepaspectratio]{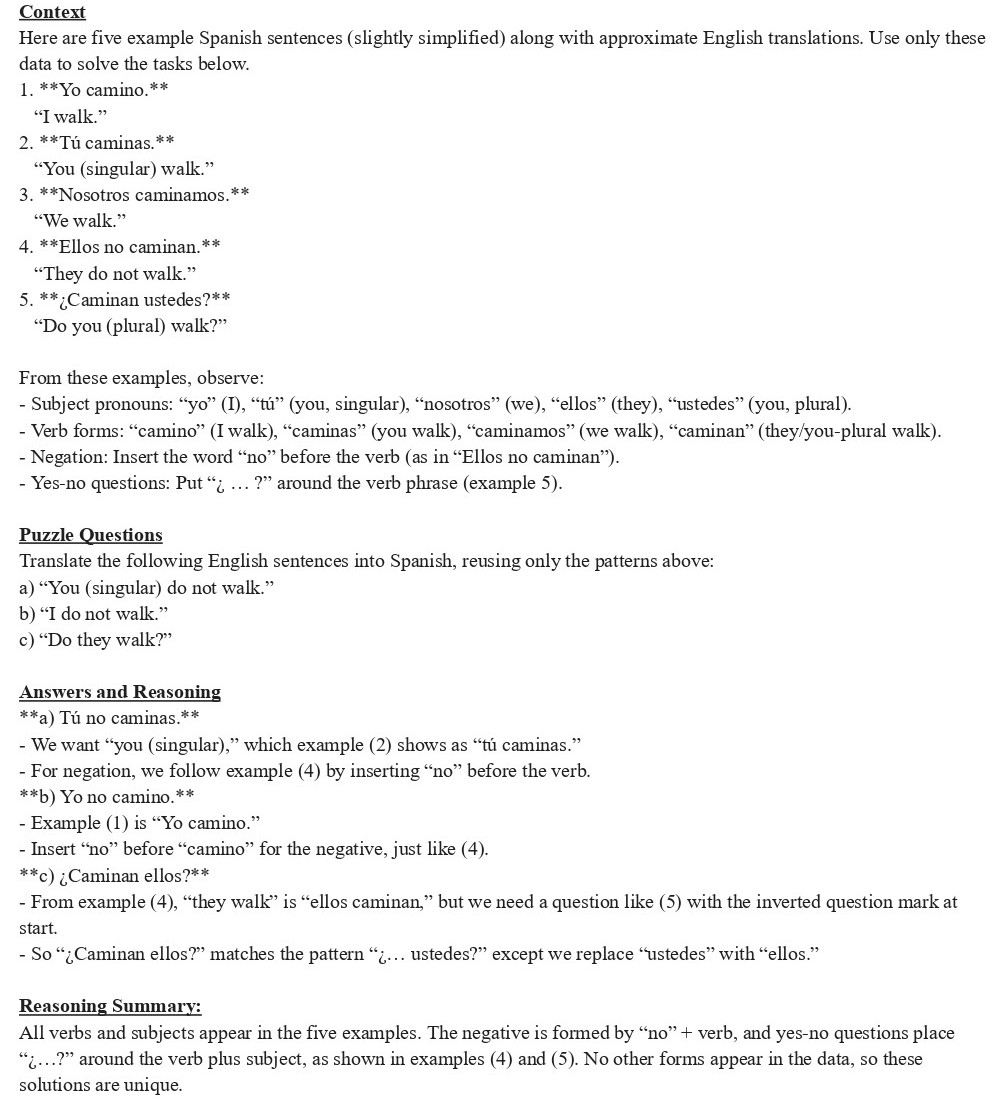}
\end{figure}
\clearpage

\subsubsection{OpenAI's o1, One-shot, Greek}
\begin{figure}[htbp]
    \centering
    \includegraphics[scale=0.8485]{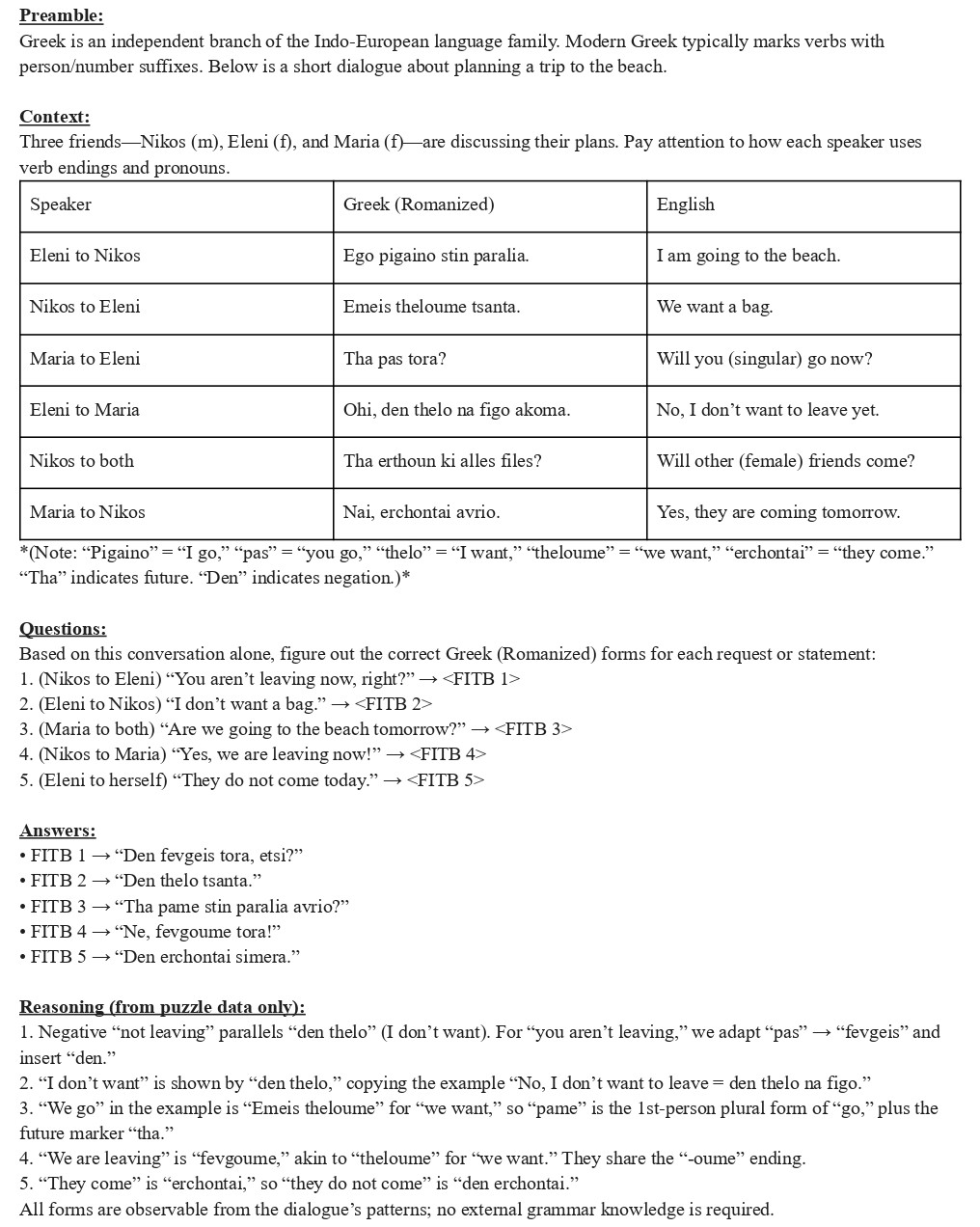}
\end{figure}
\clearpage

\subsubsection{OpenAI's o1, One-shot, Gujarati}
\begin{figure}[htbp]
    \centering
    \includegraphics[scale=0.85]{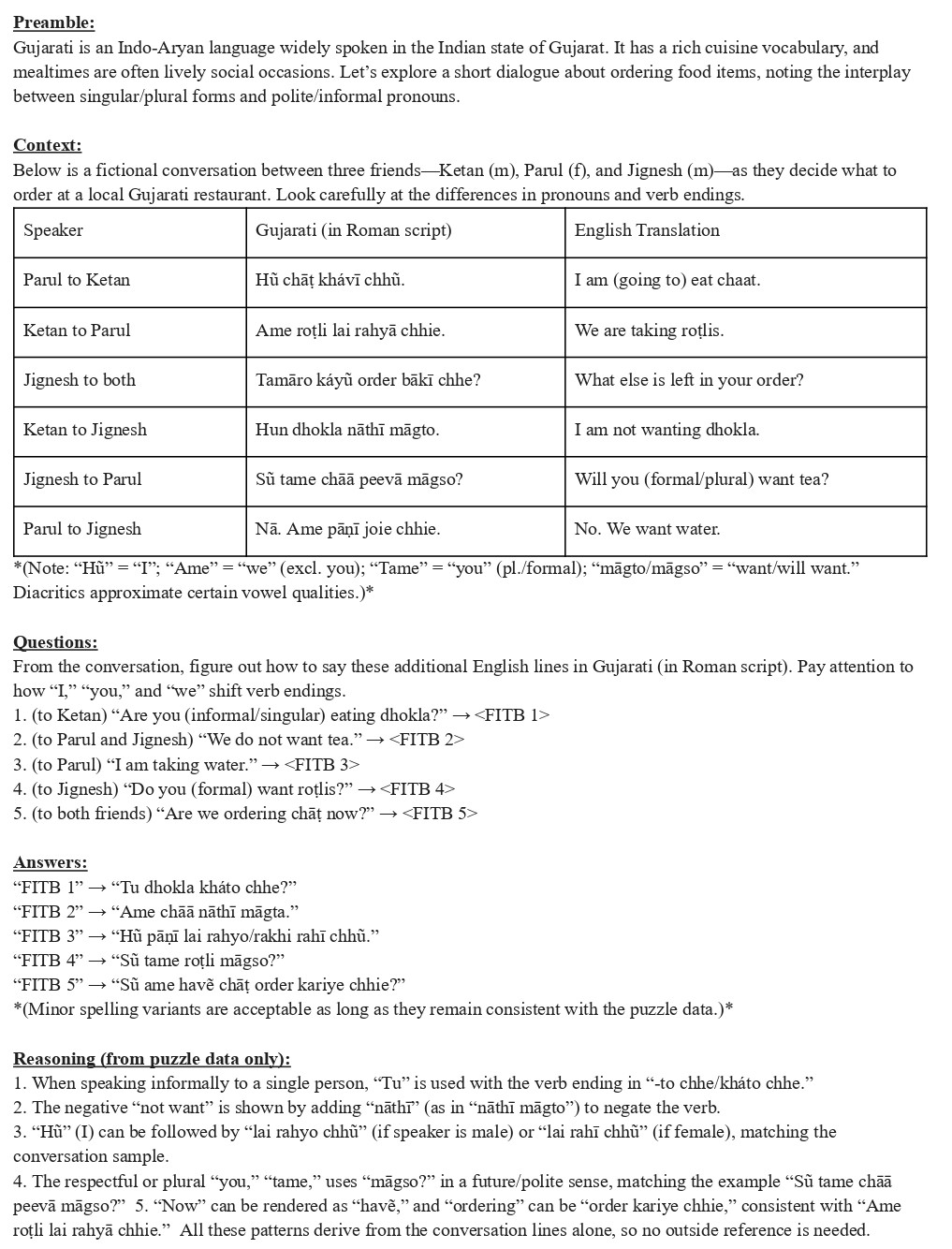}
\end{figure}
\clearpage

\subsubsection{OpenAI's o1, One-shot, Spanish}
\begin{figure}[htbp]
    \centering
    \includegraphics[scale=0.85]{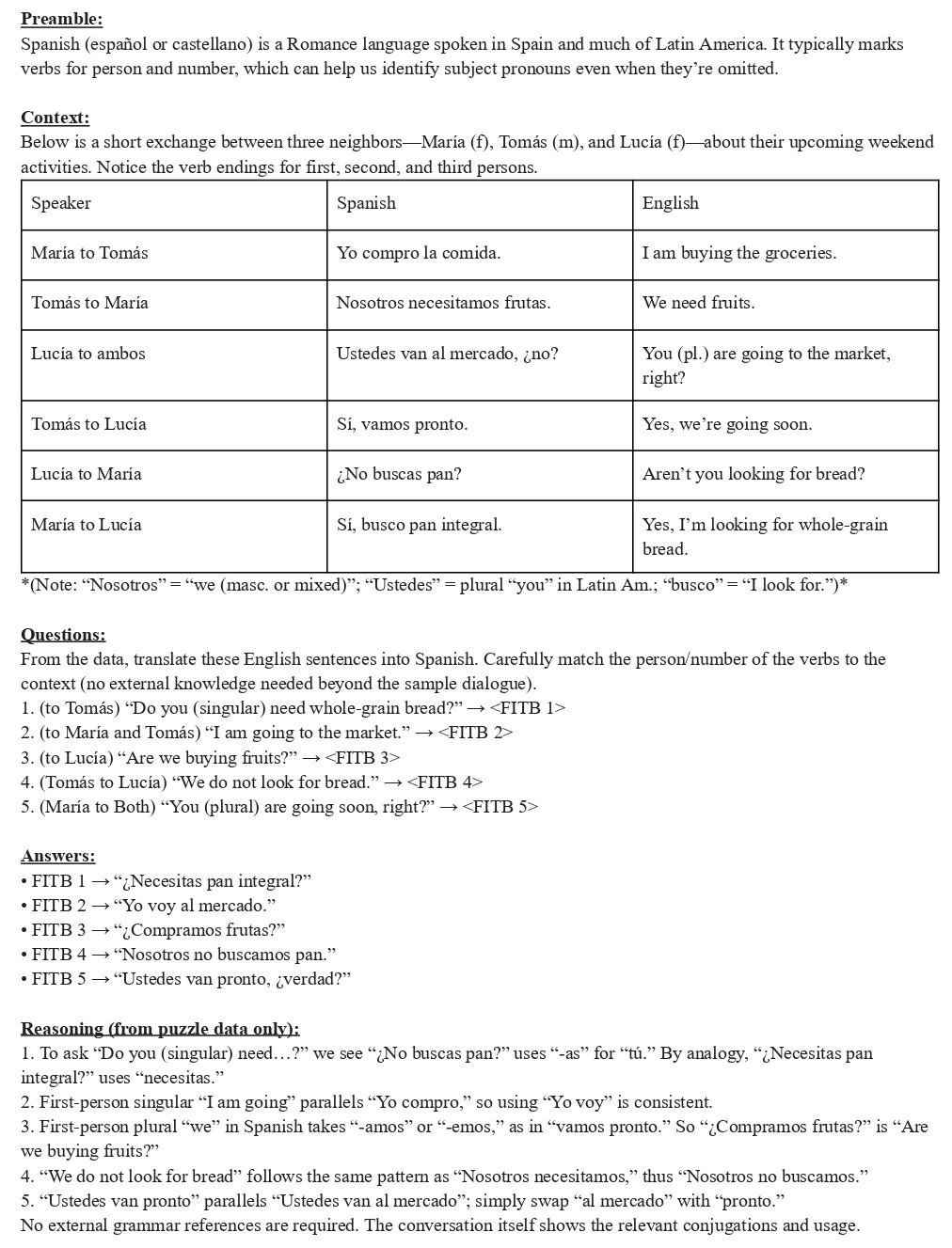}
\end{figure}
\clearpage

\subsubsection{OpenAI's o1, Few-shot, Greek}
\begin{figure}[htbp]
    \centering
    \includegraphics[scale=0.9]{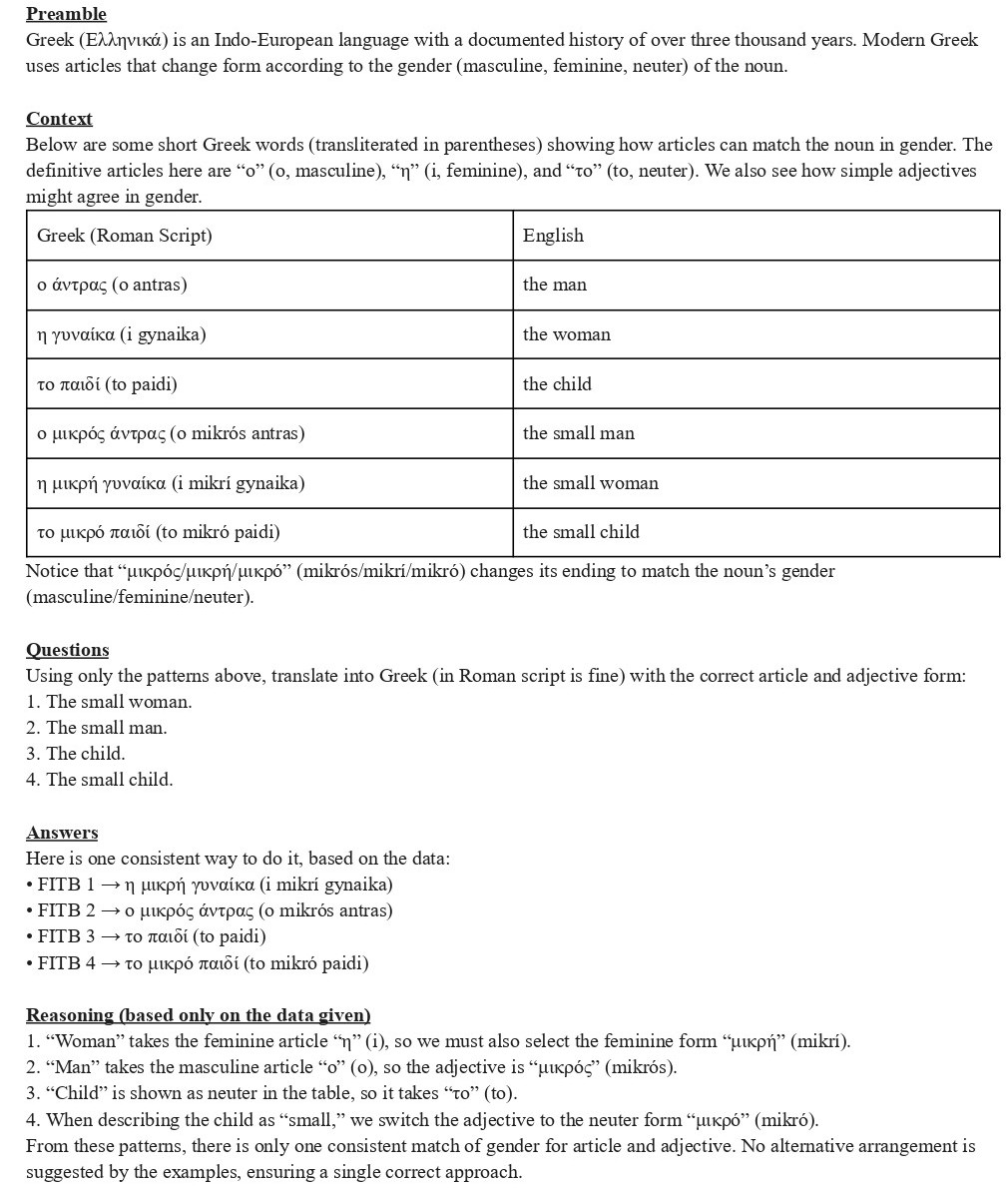}
\end{figure}
\clearpage

\subsubsection{OpenAI's o1, Few-shot, Gujarati}
\begin{figure}[htbp]
    \centering
    \includegraphics[scale=0.9]{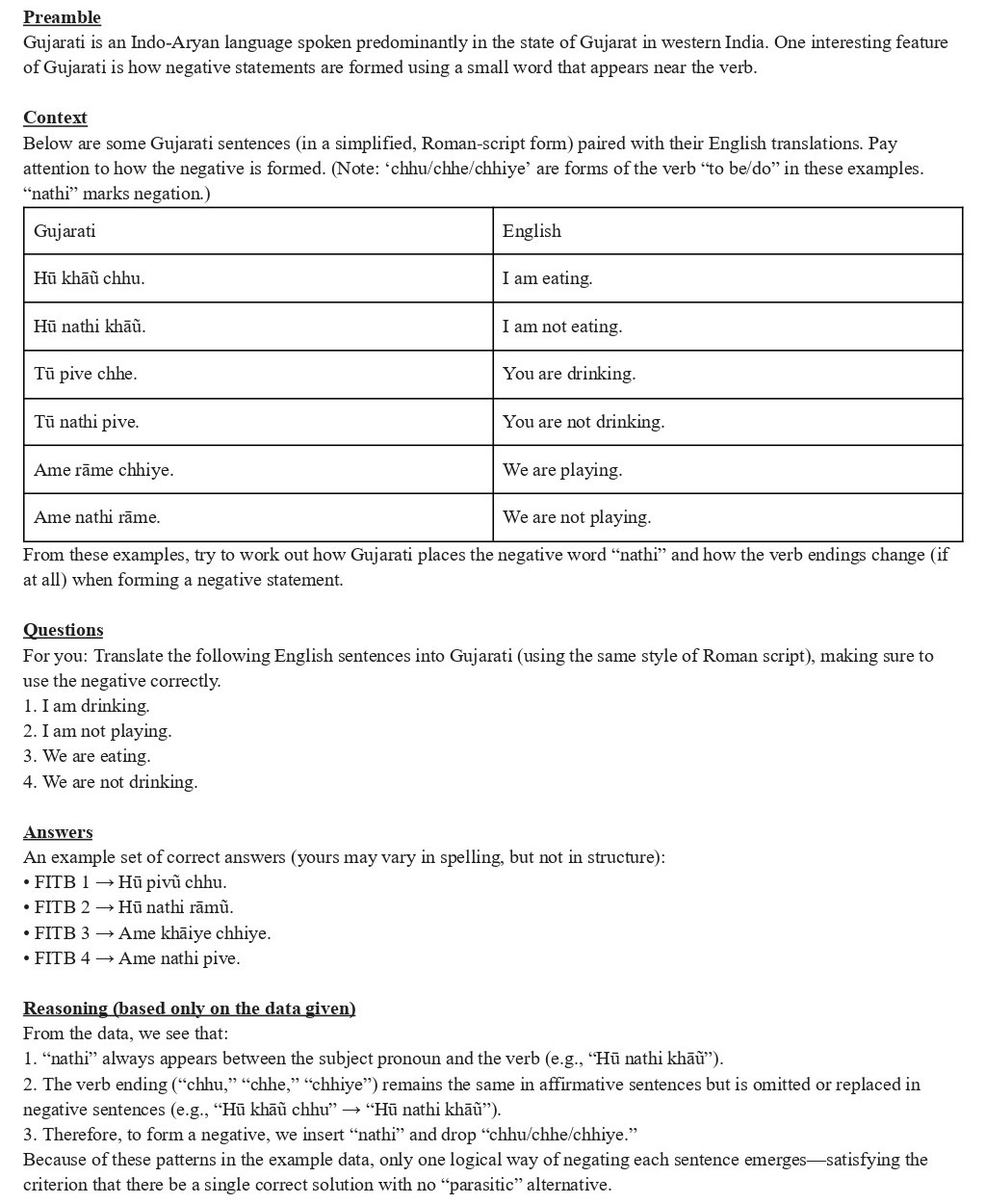}
\end{figure}
\clearpage

\subsubsection{OpenAI's o1, Few-shot, Spanish}
\begin{figure}[htbp]
    \centering
    \includegraphics[scale=0.9]{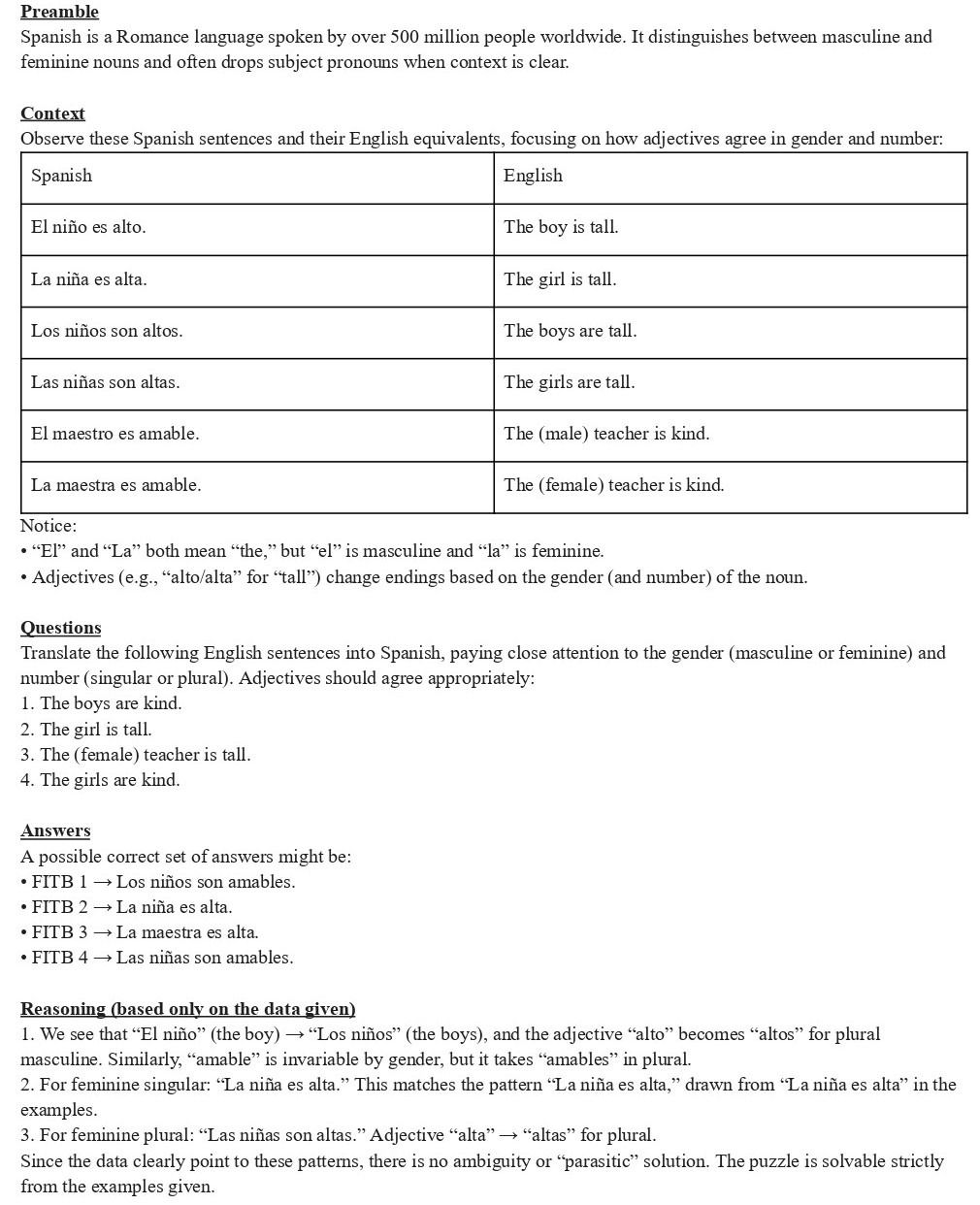}
\end{figure}